\documentclass{article}

\usepackage[preprint]{neurips_2022}

\usepackage[utf8]{inputenc} % allow utf-8 input
\usepackage[T1]{fontenc}    % use 8-bit T1 fonts
\usepackage{hyperref}       % hyperlinks
\usepackage{url}            % simple URL typesetting
\usepackage{booktabs}       % professional-quality tables
\usepackage{amsfonts}       % blackboard math symbols
\usepackage{nicefrac}       % compact symbols for 1/2, etc.
\usepackage{microtype}      % microtypography
\usepackage{xcolor}         % colors
\usepackage{algorithmic}
\usepackage{amsmath, amssymb, nccmath}
\usepackage{etoolbox}
\usepackage{placeins}
\usepackage{textcomp}
\usepackage{graphicx}
\usepackage{multicol}

\newcommand\norm[1]{\left\lVert#1\right\rVert}

\usepackage[ruled,vlined]{algorithm2e}

\title{Deep Reinforcement Learning for Multi-class Imbalanced Training}

\author{%
  Jenny Yang\textsuperscript{1} \\
  \texttt{jenny.yang@eng.ox.ac.uk} \\
   \And
   Rasheed El-Bouri \\
   \texttt{rasheedelbouri1@gmail.com} \\
   \AND
   Odhran O'Donoghue\textsuperscript{1} \\
   \texttt{odhran.odonoghue@eng.ox.ac.uk} \\
   \And
   Alexander S. Lachapelle\textsuperscript{1} \\
   \texttt{alexanderlachapelle@gmail.com} \\
   \And
   Andrew A. S. Soltan\textsuperscript{2,3} \\
   \texttt{andrew.soltan@cardiov.ox.ac.uk} \\
   \And
   David A. Clifton\textsuperscript{1} \\
   \texttt{david.clifton@eng.ox.ac.uk} \\
}

\begin{document}

\maketitle
\begin{center}
\textsuperscript{1}Institute of Biomedical Engineering, Dept. Engineering Science, University of Oxford 

\textsuperscript{2}John Radcliffe Hospital, Oxford University Hospitals NHS Foundation Trust 

\textsuperscript{3}RDM Division of Cardiovascular Medicine, University of Oxford

\end{center}

\begin{abstract}
With the rapid growth of memory and computing power, datasets are becoming increasingly complex and imbalanced. This is especially severe in the context of clinical data, where there may be one rare event for many cases in the majority class. We introduce an imbalanced classification framework, based on reinforcement learning, for training extremely imbalanced data sets, and extend it for use in multi-class settings. We combine dueling and double deep Q-learning architectures, and formulate a custom reward function and episode-training procedure, specifically with the added capability of handling multi-class imbalanced training. Using real-world clinical case studies, we demonstrate that our proposed framework outperforms current state-of-the-art imbalanced learning methods, achieving more fair and balanced classification, while also significantly improving the prediction of minority classes.
\end{abstract}

\section{Introduction}

In machine learning, predictive modeling for classification tasks involves determining the class membership of a given observation. However, in many cases, the distribution of samples across different classes is skewed. This data imbalance poses a challenge in many real-world machine learning tasks, as imbalanced classification has been shown to result in poorer predictive performance, especially in terms of the minority class (Haixiang et al., 2017; Kaur et al., 2019). Additionally, in cases where imbalance is more severe, standard machine learning techniques are often inadequate, as most of these were designed on the assumption that classes are equally distributed (Ganganwar, 2012; Zong et al., 2013). Even though skewed distributions can be learned by models if given sufficient data, there can often be similar conditional distributions between classes, which has been shown to negatively affect model performance (Denil \& Trappenberg, 2010). Thus, it is often necessary to employ specialized techniques in order to account for these data imbalances. This issue exists in many important domains, including fraud detection, anomaly detection, and disease diagnosis; and in these cases, the correct classification of the minority class is often more critical than the majority class. In this paper, we propose a general framework for training algorithms on both binary and multi-class imbalanced datasets, using deep reinforcement learning. While the framework can be applied across disciplines, we focus on clinical applications for four reasons. First, because instances of “diseased” patients generally occur much less frequently than those of “healthy” patients, making it an ideal domain to investigate our framework. Second, because the issue of class imbalance is compounded by the multi-class nature of clinical problems, and by the dozens (sometimes hundreds) of possible diagnosis codes. Third, because of inter-disease heterogeneity. And finally, because a bias towards the majority class can have severe consequences in real-world settings, where patients belonging to a minority class could receive worse care.

Two general categories of approaches have been introduced for overcoming data imbalance issues in machine learning: data-level approaches and algorithm-level approaches (Haixiang et al., 2017; Kaur et al., 2019; Tyagi \& Mittal, 2020). Data-level solutions include re-sampling the training data to make it more balanced. Two straightforward ways of doing this are: 1) adding instances of the underrepresented class or 2) removing instances from the overrepresented class. These are known as oversampling or undersampling, respectively. Both techniques can be problematic, as oversampling can increase the chances of overfitting since it utilizes exact copies of the minority class (and additionally, can lead to solutions that tend towards memorization, rather than learning a rule for the minority class), and under-sampling discards data points (wasting information that could be used for learning), making it harder for an algorithm to learn the decision boundary between classes (He \& Ma, 2013; Fernandez et al., 2018). Another oversampling approach is generating synthetic samples of the underrepresented class. A common method using this approach is the Synthetic Minority Oversampling Technique (SMOTE), which is an algorithm that selects two or more similar minority instances (using a distance measure), and perturbs an instance, one attribute at a time, by a random amount within the difference of the two chosen instances (Chawla et al., 2002). Although SMOTE generates new samples, it does not consider the majority class, which can lead to more ambiguous samples if there is significant overlap between classes. Furthermore, for clinical applications, there are not always clinical quality measures or evaluation metrics for synthetic data, making it harder to adopt into practice (Chen et al., 2021). Additionally, many clinical variables such as diagnoses are categorical, so perturbing may not be appropriate or relevant, and thus, reduces the potential variability of the generated samples. 

A common algorithm-level approach includes using a penalized/cost-sensitive model. Cost-sensitive learning directly modifies an algorithm by setting different weighted costs to adjust for the bias of each respective class. As deep models are typically trained by backpropagation of an error, errors from each class are treated equivalently, which for imbalanced training, skews a model more towards one class than another. As many machine learning methods are trained through gradient-based learning, equally weighting the losses of imbalanced classes leads to fitting parameters that are biased to the majority class (due to the aggregation of the losses prior to calculating gradients).

% In addition to data-level and algorithm level approaches, threshold adjustment has also been used to improve the balance between sensitivity and specificity during imbalanced training (Sinha et al., 2004; Chen et al., 2006). The raw output of many ML classification algorithms is a probability of class membership, which is then mapped to a particular class. For binary classification, the default threshold is typically 0.5, where values equal to or greater than 0.5 are mapped to one class and all other values are mapped to the other. However, this threshold can result in poor sensitivity, particularly when there is a large class imbalance. Thus, adjusting the decision boundary is a technique that has been used to improve false positive/false negative rates, at the time of testing.  

Reinforcement learning (RL) - where an agent learns a task by trial and error, using a reward system - has demonstrated promising results for a wide variety of tasks (Section B of the Supplementary Material). Recently, Lin et al. (2019) showed that a deep Q-network (DQN) was effective for imbalanced classification. Using both image and text datasets, they demonstrated that their method achieved better balanced classification on imbalanced datasets than other imbalanced classification methods, such as over-/under-sampling, using cost-sensitive weights, and using decision-threshold adjustment. Although they demonstrated a strong RL-based classifier, they only evaluated their method on data imbalance ratios of up to 10\%. However, in the case of clinical data (which we are focused on), there may be one rare event for hundreds or thousands of cases in the majority class. Furthermore, they only evaluated binary classification tasks; but often, a higher degree of granularity is required, as binning values to fewer (i.e. binary) classes may not be biologically relevant (especially when classes are categorical) and is heavily biased on the sample population (Yang et al., 2022a).

Thus, to extend the work of Lin et al. (2019), we formulated a framework that can be applied to multi-class classification problems, while being robust to extreme class imbalances ($\gg$ 90\%). We trained double dueling deep Q-networks (DDQN) with a specialized reward function, for the purpose of mitigating data imbalances. We evaluated our framework on two independent, real-world, imbalanced-class clinical tasks - COVID-19 screening using anonymized electronic health record (EHR) data from hospital emergency rooms (binary prediction) and patient diagnosis in ICU wards using the eICU Collaborative Research Database (eICU-CRD) (multiclass prediction). Although we use clinical case studies, our methods can be adapted to many different classification tasks. Therefore, through our study, we hope to encourage and demonstrate the effectiveness of deep reinforcement learning on a wider range of prediction tasks, including those that are multi-class in nature, and may have extreme data imbalances.

\section{The Q-imb Method}

We extend the work of Lin et al. (2019) to propose Q-imb, a framework to apply Q-learning to both binary and multi-class imbalanced classification problems. 

To formulate classification as a reinforcement learning task, we model our problem as a sequential decision-making task using a finite Markov Decision Process (MDP) framework. A typical MDP is defined using a tuple of five variables ($s$, $a$, $p$, $r$, $\gamma$), where: $s$ is the state space of the process, $a$ is the action that an agent takes, $p$ is the transition probability that results from an action, $r$ is the reward expected for a given action, and $\gamma$ is the discount rate for future rewards. 

During training, a batch of data is randomly shuffled and presented to the model in order. The features of each sample presented makes up the state $s$, and the subsequent action selected $a$ is used to select a label for classification. The reward $r$ is derived from the accuracy of classification. For a given NxD dataset, where N = number of samples, D = feature dimensionality, and K = number of classification categories; each sample  $s$, has dimensionality D; and each action $a$ is selected from one of K classes.

Here, $p$ is deterministic, as the agent moves from one state to the next according to the order of samples in the training data. Because the selection of $a$ does not determine the following sample $s$ presented to the agent, an alternative dependency must be introduced between $s$ and $a$. To this end, when an agent incorrectly classifies a minority class, the training episode is terminated, preventing any further reward $r$. This allows for a relationship between $s$ and $a$ to be learned.  

\subsection{Defining Reward for Multi-class Imbalance}

The reward, $r_t$, is the evaluation signal measuring the success of the agent’s selected action. A positive reward is given when the agent correctly classifies the sample, and a negative reward is given otherwise, thus allowing the agent to learn the optimal behavior for prediction. To accommodate for class imbalance in multi-class training data, we make the reward proportional to the relative presence of a sample in the data. The absolute reward value of a sample from the minority class is thus higher than that in the majority class, making the model more sensitive to the minority class. With $l_k$ as the label of the sample, the reward function used is: 

\noindent\begin{minipage}{.5\linewidth}
\begin{equation}
    R(s_k, a_k, l_k) =
    \begin{cases}
      \lambda_k & \text{if $a_k = l_k$}\\
      -\lambda_k  & \text{if $a_k \neq l_k$}
    \end{cases}       
  \end{equation}
\end{minipage}%
\begin{minipage}{.5\linewidth}
\begin{equation}
    \lambda_{\mathrm{k}} = \frac{\frac{1}{N_{k}}}{\norm{\frac{1}{N_{0}}, \frac{1}{N_{1}}, ..., \frac{1}{N_{k}} }^2} 
  \end{equation}
\end{minipage}

$N_k$ represents the number of class instances in class $k$ and $\lambda$ is a trade-off parameter used for adjusting the influence of the minority and majority classes. We found that our model achieved desirable performance when $\lambda$ is the vector-normalized reciprocal of the number of class instances, as shown in Eq. 2. 

To balance immediate and future rewards, a discount factor, $\gamma \epsilon [0,1)$, is used.  

% The behavior learned by the agent is known as a deterministic policy, where $\pi$ is the mapping function of obtained inputs to actions, $\pi_{\theta}(a|s)$, and $\theta$ are the parameters to be learned. Thus, the aim of training is to define the classification task by an optimal classification policy, $\pi$*. As we are training a finite MDP, it is known that there exists an optimal policy (i.e., a policy that maximizes the expected cumulative reward from the initial state).  

\subsection{Algorithmic Adaptations for Multi-class Imbalance}

To effectively perform reinforcement learning under a multi-class imbalance setting, we propose the use of a double dueling deep Q-network (D-DDQN). We first begin by discussing the Q-value function. We then proceed to justify the use of dueling and double Q-learning components in the domain of class-imbalanced multi-class learning.

% \section{Methods}

% \subsection{Markov Decision Process}

% To model our problem as a sequential decision-making task, we use a finite Markov Decision Process (MDP) framework. We define our MDP using a tuple of five variables ($s$, $a$, $p$, $r$, $\gamma$), where: $s$ is the state space of the process, $a$ is the action that an agent takes, $p$ is the transition probability that results from an action, $r$ is the reward expected for a given action, and $\gamma$ is the discount rate for future rewards. We train our RL framework in finite episodes (training sections), where each episode is a sequence of states, actions, and rewards. The training data is randomly shuffled at the beginning of each new episode.

% \textbf{State, \textit{s}:} At each time step, $t$, the state, $s_t$, represents the state space of the current sample, $x_t$, and any already selected samples. 

% %%Figure out how to change the type of 'a'
% \textbf{Action, \textit{a}:} The action, $a_t$, is the prediction the agent makes at $s_t$, with respect to $x_t$. Here, $a$ = $\{0,1\}$, where 0 corresponds COVID-19 negative cases and 1 corresponds to COVID-19 positive cases.

% \textbf{Transition Probability, \textit{p}:} The transition probability, $p(s_{t+1}|s_t, a_t)$ is the probability of transitioning to state $s_{t+1}$ when selecting the action $a_t$ in state $s_t$. Here, $p$ is deterministic, as the agent moves from one state to the next according to the order of samples in the training data.

\subsubsection{Policy Iteration by Q-Learning}

% The value of a state-action combination under the stochastic policy $\pi$, $v^\pi(s_t)$, is the expected cumulative reward for a process starting in the initial state $s$. This is calculated using the function: 

% \noindent\begin{minipage}{.5\linewidth}
% \begin{equation}
%     V^\pi(s) = E[g_t|s_0 = s],
% \end{equation}
% \end{minipage}%
% \begin{minipage}{.5\linewidth}
% \begin{equation}
%     g_t = \sum_{t=0}^{\infty}\gamma^tR(s_t,\pi(s_t)),
% \end{equation}
% \end{minipage}

% where $g_t$ is the cumulative reward from all decisions at all steps.

An optimal policy $\pi$* is a policy that maximizes $v^\pi$ (the value of a state-action combination), and this can be discovered by iterating through a series of policies, $\{\pi\}_i^k$, where $\pi$*=$\pi^k$. Using the Bellman equation, $v^\pi$ can be solved for by solving a system of linear equations and calculating a set of $Q$-values, where $Q$ represents the action-value function: 

\begin{equation}
    Q_i^\pi(s_t,a_t)=r(s_t, a_t)+\gamma\sum_{s_{t+1}}p(s_{t+1}|s_t, a_t)v_i^\pi( s_{t+1}),
\end{equation}
which gives successive policies:

\begin{equation}
    \pi_{i+1}(a_t|s_t)=arg \max_aQ_i^\pi(s_t, a_t),
\end{equation}

where, $\pi^* = arg\max_aQ^*$. Finally, to relate the state-action value function and $Q$ function, we can use the advantage function:

% \begin{equation}
%     \pi^*(a_t|s_t)=arg\max_aQ^*(s_t, a_t),
% \end{equation}

\begin{equation}
    A^\pi(s_t,a_t)=Q^\pi(s_t,a_t)-V^\pi(s_t)
\end{equation}

%The value function, $V^\pi$ can be viewed as a proxy for the "goodness" of a particular state, and the $Q^\pi$ function evaluates the value of selecting a particular action in this state (Wang et al., 2016). Thus, $A^\pi$ can be interpreted as the relative importance of each action.  

\subsubsection{Dueling Q-Network Architecture}

In the standard DQN setup, the output layer of the network corresponds to predicted Q-values for state-action pairs. In situations with a high number of possible state-action pairs such as multi-class prediction tasks, it becomes difficult to provide update information about the state because only one state-action pair in a state can be trained at a time. To alleviate this, the Dueling DQN provides a method to train state representations independently of action representations. 

For a DQN, the Q-network implemented is a single-stream neural network. This is a standard neural network with a continuous sequence of fully connected layers. The Dueling Q-network instead implements a fully-connected neural network with two streams - one to estimate the value (scalar) and another to estimate the advantages of each action (vector). These two streams are combined to produce a single output, which is the $Q$ function. This is known as a dueling network (Wang et al., 2016), and is shown in Section C of the Supplementary Material.

Based on the definition of the advantage function, we represent $Q$ as:

\begin{equation}
        Q_i^\pi(s_t,a_t;\theta_i,\alpha_i, \beta_i) =V_i^\pi(s_t;\theta_i,\beta_i)+
        \bigg(A_i^\pi(s_t,a_t;\theta_i,\alpha_i)-softmax(A_i^\pi(s_t,a_{t+1};\theta_i,\alpha_i))\bigg),
\end{equation}

where $\alpha$ and $\beta$ represent the parameters of the $A$ and $V$ streams of the fully connected layers, respectively.
The additional softmax module is to allow $Q$ to recover $V$ and $A$ uniquely. Additionally, this extra term does not change the relative rank of $A$ (and subsequently, Q-values), which preserves the $\epsilon$-greedy policy (which we use in our training). A full explanation for adding this additional module can be found in Wang et al. (2016). 

\begin{figure}[ht]
\centering
\includegraphics[scale=0.4]{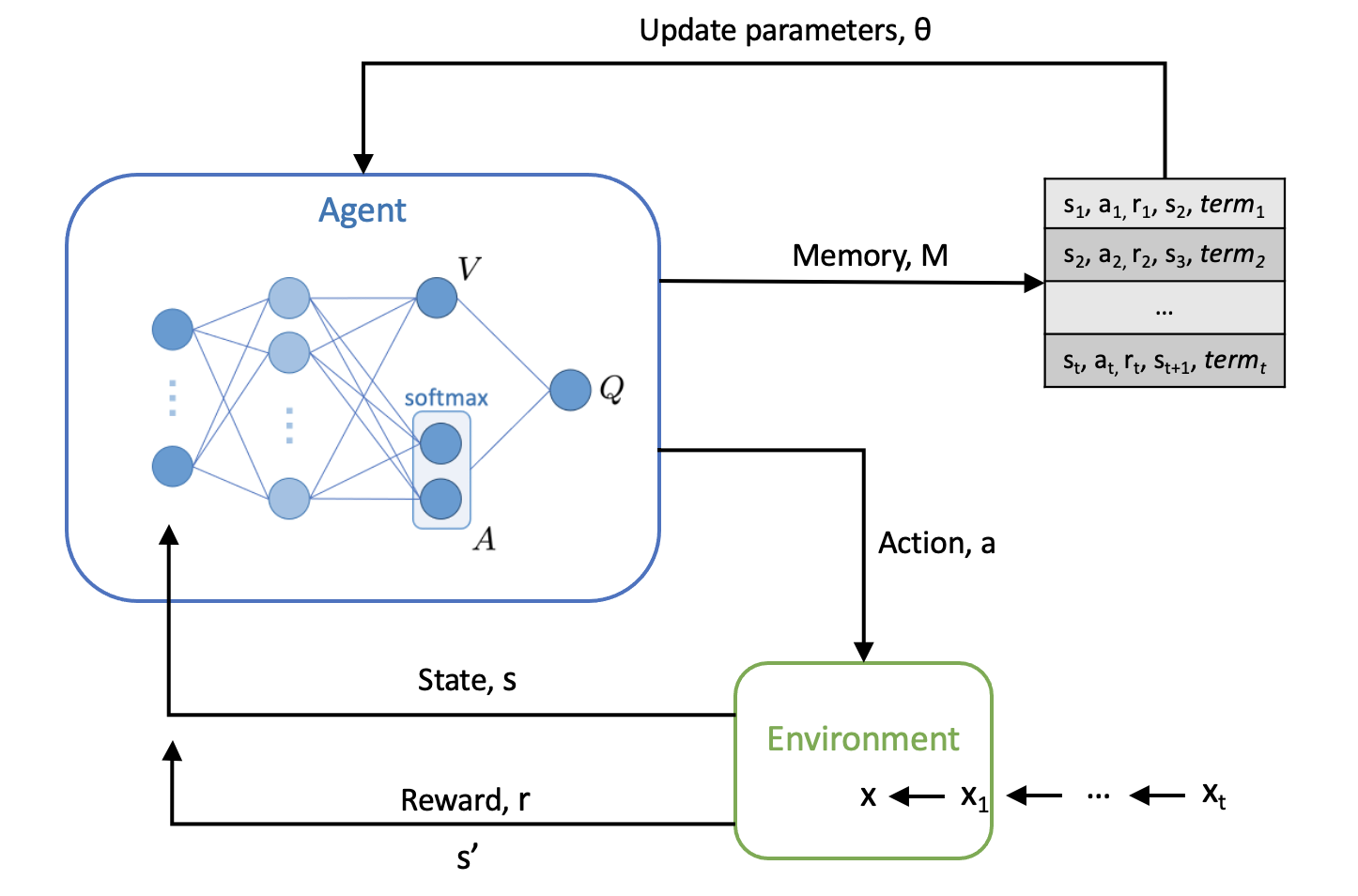}
\caption{Overview of the reinforcement learning framework used. A dueling network architecture, with two streams to independently estimate the state-values (scalar) and advantages (vector) for each action, is
shown.}
%\label{fig:dueling}
\end{figure}

\subsubsection{Double Deep Q-Learning}
In each episode, the combinations of states, actions, and rewards at each step, $(s_t, a_t, r_t, s_{t+1})$, are stored in the agent’s working memory, M. To learn the parameters of the $Q$-network $\theta$, a randomly sampled subset of these transitions $B$, are used in the gradient descent step. The $Q$-network is optimized using the mean-squared error loss function:

\begin{equation}
    L(\theta_i)=\sum_{(s_t,a_t,r_t,s_{t+1})\epsilon B}(y-Q(s_t,a_t; \theta_i))^2
\end{equation}
As in traditional supervised learning, $y$ can be viewed as the target to be predicted and $Q(s,a; \theta_i)$ is the prediction. We implement $y$ following the format of a Double Deep Q-Network (DDQN).

It has been shown that a standard DQN is more likely to give overoptimistic value estimates for actions (Thrun and Schwartz, 1993). A DQN uses the current Q-network to determine an action, as well as estimate its value. This increases the likelihood of selecting overestimated values (from the maximization step), making it harder to learn the optimal policy, as overestimations can occur even when action values are incorrect. This can become even more complex when there are a high number of possible state-action pairs, as seen in multi-class scenarios. Double Deep Q-Learning (DDQN) was introduced as a method of reducing this overestimation (van Hasselt et al., 2015). Unlike a DQN, a DDQN, decouples the selection and evaluation steps, and instead, uses the current Q-network to select actions, and the target Q-network to estimate its value (Sui et al., 2018). Thus, a separate set of weights, $\theta'$, are used to provide an unbiased estimate of value.

We implement the DDQN algorithm using the following target function: 

\begin{equation}
    y_i= r_t+(1-term)\gamma Q(s_{t+1}, arg\max_a Q(s_{t+1},a_{t+1};\theta_i); \theta_{i}')
\end{equation}

As previously mentioned, a dependency between a state and action needs to be established for the agent to learn a relationship. Thus, within this function, the value of $term$ is set to 1 once the agent reaches its terminal state, and 0 otherwise. A terminal state is reached after the agent has iterated through all samples in the training data (or a set number of samples, specified at the beginning of training), or when the agent misclassifies a sample from the minority class(es) (preventing any further reward). 

\section{Model Comparators and Evaluation Metrics}

We compare the effectiveness of Q-imb against three baseline models - a fully-connected neural network, XGBoost, and a DDQN with no dueling component (each with no added imbalanced learning strategy applied). We also compare results of models with the addition of two commonly used, state-of-the-art imbalanced data-learning methods:

\textbf{SMOTE}: SMOTE was applied to the training set using a minority oversampling strategy of 0.2 (i.e. the minority class was oversampled to have 20\% of the number of samples in the majority class).

\textbf{Cost-Sensitive Learning}: Different weighted costs were assigned to each class during training. The value of class weights chosen were inversely proportional to class frequencies in the training data.

We trained both a neural network and XGBoost model as-is, and additionally trained implementations that utilized SMOTE and cost-sensitive learning. Appropriate hyperparameter values for all models were determined through standard 5-fold cross-validation (CV), using the training set. For the DDQN without a dueling component, we use the same hyperparameter settings as Q-imb, to directly compare balanced classification performance of both methods. Details on network architecture and final hyperparameter values used for each model can be found in Section H of the Supplementary Material. 

To evaluate the classification performance, we calculate the sensitivity, specificity, and the area under receiver operator characteristic curve (AUROC) across all test sets.

As the purpose of our model is to effectively train models on imbalanced data, we use F-measure and G-mean metrics to evaluate the balanced classification performance of each model. These are known as the harmonic and geometric means, respectively. A higher F value implies that sensitivity and precision are simultaneously high (Gu et al., 2009). Similarly, the geometric mean evaluates sensitivity and specificity, ensuring that both the true positive and true negative rates are high (Gu et al., 2009). Equations for F and G metrics can be found in Section E of the Supplementary Material.

\section{Prediction Tasks and Datasets}

\textbf{COVID-19 Status Prediction:} To show that Q-imb is effective for binary classification, we train models to predict the COVID-19 status for patients presenting to hospital emergency departments across four United Kingdom (UK) National Health Service (NHS) Trusts, using anonymized EHR data (specifically, blood tests and vital sign features). We trained and optimized our model using 114,957 COVID-free patient presentations from OUH prior to the global COVID-19 outbreak, and 701 patient presentations during the first wave of the COVID-19 epidemic in the UK that had a positive PCR test for COVID-19 (ensuring that the label of COVID-19 status was correct during training). We then performed validation on a prospective OUH cohort, as well as external validation on three additional patient cohorts from PUH, UHB, and BH (totalling 72,223 admitted patients, including 4,600 of which were COVID-19 positive). During training, we used a simulated disease prevalence of 5\% (i.e. a data imbalance ratio of 1 positive COVID-19 case: 20 negative controls). This aligns with real COVID-19 prevalences at all four sites (during the dates of data extraction), which ranged between 4.27\%-12.2\%.

\textbf{Patient Diagnosis:} Further analysis was performed using the eICU Collaborative Research Database (eICU-CRD) (Pollard et al., 2018). Using this database, we tried to predict which of five acute events (cardiovascular, respiratory, gastrointestinal, systemic, renal) a patient is diagnosed with during their ICU stay. Through this task, we aimed to evaluate the utility of our model for a multi-class task, which is often necessary for many real-world applications. Prevalences across events ranged from 8.7\%-33.6\%, making this an appropriate task for investigating data imbalance effects. We trained our model using 18,076 samples and then tested it on 6,026 held-out samples. 

The full inclusion and exclusion criteria for patient cohorts, summary population statistics, features used in training, and data pre-processing steps for both tasks can be found in Sections F and G of the Supplementary Material, respectively.

\section{Results}

\subsection{COVID-19 Diagnosis}

Table 1 shows results for COVID-19 prediction, where we performed prospective validation and external validation across four NHS Trusts. Scores for F-measure and G-mean are presented alongside sensitivity, specificity, and AUROC (with 95\% CIs). The results presented use an adjusted decision threshold, optimized to a sensitivity of 0.9. This threshold was chosen to ensure clinically acceptable performance in detecting positive COVID-19 cases, while also exceeding the sensitivities of current diagnostic testing methods (lateral flow device sensitivity is around 57\% [Soltan et al., 2022], real-time polymerase chain reaction sensitivity is about 70\% [Goudouris, 2021]). Results achieved with no threshold adjustment (i.e., using the default threshold of 0.5) and when using a threshold of 0.85 can be found in Section I of the Supplementary Material. As we are focused on evaluating balanced classification, we use red and blue to depict the best and second best scores, respectively, for F and G. 

In terms of balanced classification, Q-imb achieved the highest F and G scores for three test sets - OUH, UHB, and BH. The XGBoost models using SMOTE and cost-sensitive weights achieved the best F and G scores on the PUH dataset. Similar results were found for models optimized to sensitivities of 0.85, with Q-imb generally achieving the highest (or second highest) F and G scores, demonstrating model consistency. When no threshold adjustment was applied, Q-imb also achieved the highest (or second highest) G scores on all test sets; however, F scores were not as high compared to other models that had much lower sensitivity ($<$0.61), but very high specificity ($>$0.93), due to the nature of how F is calculated. The non-dueling DDQN model consistently achieved the lowest F and G scores, across all test sets. Comparison of the output from Q-imb to all other methods was found to be statistically significant (p$<$0.0001, by the Wilcoxon Signed Rank Test).

All models trained achieved reasonably high AUROC scores across all test sets, comparable to those reported in previous studies, using the same patient cohorts (Soltan et al., 2022; Yang et al., 2022a, Yang et al., 2022b) (Section F of the Supplementary Material), demonstrating that Q-imb is a strong classifier, in addition to being able to account for data imbalances.

\begin{table}[ht]
\caption{Performance metrics for COVID-19 prediction. Results reported as F-measure, G-mean, AUROC, sensitivity, and specificity for OUH, PUH, UHB, and BH test sets; 95\% confidence intervals (CIs) also shown. Red and blue values denote best and second best scores, respectively, for F-measure and G-mean. Threshold adjustment applied to optimize models to sensitivities of 0.9.}
\resizebox{\textwidth}{!}{%
\begin{tabular}{@{}llllll@{}}
\toprule
Model                           & F                                     & G                                     & AUROC               & Sensitivity         & Specificity         \\ \midrule
\multicolumn{6}{l}{\textbf{OUH}}                                                                                                                                                  \\
Reinforcement Learning (Q-imb) & {\color[HTML]{CB0000} \textbf{0.426}} & {\color[HTML]{CB0000} \textbf{0.770}}  & 0.861 (0.850-0.871) & 0.838 (0.822-0.854) & 0.707 (0.701-0.713) \\
Reinforcement Learning (DDQN)   & 0.306                                 & 0.534                                 & 0.758 (0.745-0.771) & 0.852 (0.836-0.867) & 0.334 (0.328-0.341) \\
Neural Network                  & 0.388                                 & 0.715                                 & 0.877 (0.867-0.886) & 0.899 (0.885-0.912) & 0.568 (0.562-0.575) \\
Neural Network + SMOTE          & 0.398                                 & 0.736                                 & 0.871 (0.861-0.881) & 0.871 (0.856-0.885) & 0.622 (0.615-0.628) \\
Neural Network + Cost-Sensitive & 0.400                                 & 0.737                                 & 0.872 (0.862-0.882) & 0.881 (0.867-0.895) & 0.616 (0.609-0.623) \\
XGBoost                         & 0.399                                 & 0.734                                 & 0.877 (0.867-0.887) & 0.889 (0.875-0.902) & 0.607 (0.600-0.614) \\
XGBoost + SMOTE                 & {\color[HTML]{3166FF} \textbf{0.422}} & {\color[HTML]{3166FF} \textbf{0.766}} & 0.876 (0.866-0.886) & 0.846 (0.830-0.862) & 0.694 (0.687-0.700) \\
XGBoost + Cost-Sensitive        & 0.399                                 & 0.739                                 & 0.869 (0.859-0.879) & 0.857 (0.842-0.872) & 0.637 (0.630-0.643) \\ \midrule
\multicolumn{6}{l}{\textbf{PUH}}                                                                                                                                                  \\
Reinforcement Learning (Q-imb) & 0.306                                 & 0.727                                 & 0.831 (0.819-0.842) & 0.828 (0.812-0.845) & 0.638 (0.633-0.643) \\
Reinforcement Learning (DDQN)   & 0.248                                 & 0.606                                 & 0.762 (0.750-0.774) & 0.804 (0.787-0.821) & 0.457 (0.451-0.462) \\
Neural Network                  & 0.289                                 & 0.676                                 & 0.857 (0.847-0.868) & 0.903 (0.890-0.916) & 0.506 (0.501-0.511) \\
Neural Network + SMOTE          & 0.309                                 & 0.728                                 & 0.856 (0.845-0.866) & 0.859 (0.844-0.875) & 0.617 (0.612-0.622) \\
Neural Network + Cost-Sensitive & 0.288                                 & 0.681                                 & 0.850 (0.839-0.861) & 0.883 (0.869-0.897) & 0.526 (0.521-0.531) \\
XGBoost                         & 0.321                                 & 0.741                                 & 0.881 (0.871-0.891) & 0.898 (0.884-0.911) & 0.612 (0.607-0.617) \\
XGBoost + SMOTE                 & {\color[HTML]{3166FF} \textbf{0.325}} & {\color[HTML]{3166FF} \textbf{0.750}} & 0.881 (0.871-0.890) & 0.877 (0.863-0.892) & 0.641 (0.636-0.646) \\
XGBoost + Cost-Sensitive        & {\color[HTML]{CB0000} \textbf{0.336}} & {\color[HTML]{CB0000} \textbf{0.766}} & 0.881 (0.871-0.891) & 0.862 (0.847-0.877) & 0.680 (0.675-0.684) \\ \midrule
\multicolumn{6}{l}{\textbf{UHB}}                                                                                                                                                  \\
Reinforcement Learning (Q-imb) & {\color[HTML]{CB0000} \textbf{0.304}} & {\color[HTML]{CB0000} \textbf{0.764}} & 0.837 (0.814-0.861) & 0.815 (0.779-0.852) & 0.717 (0.708-0.726) \\
Reinforcement Learning (DDQN)   & 0.209                                 & 0.516                                 & 0.721 (0.694-0.749) & 0.841 (0.806-0.875) & 0.317 (0.308-0.326) \\
Neural Network                  & 0.279                                 & 0.718                                 & 0.866 (0.844-0.888) & 0.913 (0.887-0.940) & 0.565 (0.555-0.574) \\
Neural Network + SMOTE          & 0.290                                  & 0.746                                 & 0.850 (0.828-0.873) & 0.845 (0.811-0.879) & 0.658 (0.649-0.668) \\
Neural Network + Cost-Sensitive & 0.284                                 & 0.733                                 & 0.861 (0.839-0.883) & 0.879 (0.849-0.910) & 0.611 (0.601-0.621) \\
XGBoost                         & 0.287                                 & 0.740                                 & 0.861 (0.839-0.883) & 0.872 (0.841-0.904) & 0.627 (0.618-0.637) \\
XGBoost + SMOTE                 & {\color[HTML]{3166FF} \textbf{0.292}} & {\color[HTML]{3166FF} \textbf{0.750}} & 0.853 (0.830-0.876) & 0.827 (0.791-0.862) & 0.680 (0.671-0.690) \\
XGBoost + Cost-Sensitive        & 0.289                                 & 0.746                                 & 0.851 (0.829-0.874) & 0.838 (0.804-0.873) & 0.663 (0.654-0.673) \\ \midrule
\multicolumn{6}{l}{\textbf{BH}}                                                                                                                                                   \\
Reinforcement Learning (Q-imb) & {\color[HTML]{CB0000} \textbf{0.561}} & {\color[HTML]{CB0000} \textbf{0.815}} & 0.867 (0.829-0.906) & 0.806 (0.741-0.870) & 0.825 (0.802-0.848) \\
Reinforcement Learning (DDQN)   & 0.362                                 & 0.589                                 & 0.706 (0.656-0.756) & 0.799 (0.733-0.864) & 0.434 (0.403-0.464) \\
Neural Network                  & 0.525                                 & 0.802                                 & 0.885 (0.849-0.921) & 0.868 (0.813-0.923) & 0.741 (0.714-0.767) \\
Neural Network + SMOTE          & {\color[HTML]{3166FF} \textbf{0.540}} & 0.801                                 & 0.882 (0.845-0.919) & 0.792 (0.725-0.858) & 0.810 (0.786-0.834) \\
Neural Network + Cost-Sensitive & 0.529                                 & {\color[HTML]{3166FF} \textbf{0.804}} & 0.883 (0.847-0.920) & 0.854 (0.797-0.912) & 0.756 (0.730-0.782) \\
XGBoost                         & 0.501                                 & 0.780                                 & 0.894 (0.859-0.929) & 0.896 (0.846-0.946) & 0.679 (0.650-0.707) \\
XGBoost + SMOTE                 & 0.535                                 & 0.803                                 & 0.885 (0.849-0.921) & 0.819 (0.757-0.882) & 0.787 (0.762-0.812) \\
XGBoost + Cost-Sensitive        & 0.511                                 & 0.790                                 & 0.889 (0.854-0.925) & 0.861 (0.805-0.918) & 0.724 (0.697-0.751) \\ \bottomrule
\end{tabular}%
}
\end{table}

\FloatBarrier

\subsection{Patient Diagnosis Prediction}

For multiclass patient diagnosis, we calculated the individual sensitivities and G-means for all classes, using a "one-vs-all" method, for each method used, and present the mean sensitivities and G-means across all classes (Table 2).

\begin{table}[ht]
\begin{center}
\caption{Performance metrics for multi-class diagnosis prediction. Results reported as average sensitivities and G values across all classes, shown alongside standard deviation.}
\scalebox{0.8}{
\begin{tabular}{@{}ccc@{}}
\toprule
Model & G & Sensitivity \\ \midrule
Reinforcement Learning (Q-imb) & {\color[HTML]{CB0000} \textbf{0.834 (0.082)}} & 0.748 (0.126) \\
Reinforcement Learning (DDQN) & 0.717 (0.139) & 0.601 (0.215) \\
Neural Network & 0.806 (0.105) & 0.715 (0.183) \\
Neural Network + SMOTE & 0.804 (0.109) & 0.714 (0.193) \\
Neural Network +   Cost-Sensitive & 0.801 (0.103) & 0.712 (0.190) \\
XGBoost & 0.819 (0.106) & 0.733 (0.181) \\
XGBoost + SMOTE & 0.819 (0.107) & 0.733 (0.184) \\
XGBoost + Cost-Sensitive & {\color[HTML]{3166FF} \textbf{0.830 (0.092)}} &  0.744 (0.142) \\ \bottomrule
\end{tabular}
}
\end{center}
\end{table}

Here, the highest mean sensitivity was achieved by Q-imb, followed closely by the XGBoost model with cost-sensitive weights. Comparison of the output from Q-imb to all other methods was found to be statistically significant (p$<$0.0001, by the Wilcoxon Signed Rank Test).

There was a wide range in sensitivities for each acute event category across all models, varying from sensitivities of $<$0.5 to $>$0.9 (individual class performances can be found in Section I of the Supplementary Material). Q-imb displayed the lowest variance (by standard deviation) for sensitivities and G-means across all classes (SDs of 0.126 and 0.082 for sensitivity and G-mean, respectively), suggesting that it achieves the most fair classification (i.e. is the least biased classification towards majority classes). This is closely followed by the XGBoost model with cost-sensitive weights (SDs of 0.142 and 0.092 for sensitivity and G-mean, respectively). As seen in the previous task, the non-dueling DDQN achieved the lowest mean sensitivity and mean G score, confirming how a policy that leads to good performance is harder to learn when state-action pairs are coupled. Q-imb also achieved either the highest (or second highest) sensitivities on the three classes with the lowest prevalences (acute gastrointestinal events, acute systemic events, acute renal events), across all methods. However, although sensitivities of minority classes improved, sensitivities for majority classes were slightly lower compared to other models. This was also the case for the XGBoost model with cost-sensitive weights, as the predictive performances of minority classes also improved at the cost of majority class prediction. However, in both cases, the absolute rate of improvement was higher, leading to better overall performance. This was also reflected by the G scores, as Q-imb achieved the highest mean G score, followed by the XGBoost model with cost-sensitive weights, suggesting more fair/balanced classification.

\section{Discussion and Conclusion}

As seen in many real-world ML studies, data imbalance poses a challenge, particularly with respect to correct classification of the minority class. This is especially evident in healthcare-related tasks, where the classification of the minority class is often more critical than the majority class. In this study, we used deep reinforcement learning, specifically in the context of imbalanced classification, and introduced a new formulation for multi-class settings. We evaluated our method against state-of-the-art imbalanced learning methods, using two challenging, real-world clinical case studies of COVID-19 diagnosis (binary) and general patient diagnosis (multi-class), with extreme data imbalances.

Experiments showed that our model achieved fairer and more balanced classification on imbalanced data than other imbalanced classification methods, significantly improving minority class sensitivity, while still achieving relatively high majority-class performance. We also demonstrated that a dueling architecture was able to learn the state-value function, and therefore, the policy, more efficiently than a non-dueling comparator. Additionally, we showed that trained models were generalizable across four out-of-sample validation data sets (from four independent hospital trusts) with varying disease prevalences/imbalance ratios, which is a common disparity between hospitals and populations. 

Although we were able to demonstrate that our method was effective for training imbalanced multi-class problems, it is still important to consider whether a class-specific model (i.e., using multiple one-vs-rest models) or a more general multi-class model is best suited for the task. Thus, future work should also consider the properties and nature of samples in minority classes, as this can give insight into the source of learning difficulties. For example, if there are different degrees of overlapping distributions between classes, it may be better to train multiple binary classifiers instead of a single multi-class one. Contrarily, for tasks where very large models need to be used, training multiple one-vs-rest models can overwhelm computing power; thus, using Q-imb in those scenarios would be beneficial. 

With respect to the binary classification task, we used threshold adjustment to ensure models achieved high sensitivity for detecting COVID-19. Although threshold adjustment can be an effective strategy for achieving desirable detection rates, it is biased on the particular dataset the threshold is determined on. It has previously been shown that data can be biased towards site-specific factors (e.g. annotation, measuring devices, collection/processing methods, cohort distributions) (Yang et al., 2021a); thus, the threshold used at one site, may not be appropriate for use at a different site with varying distributions. This can make it difficult to perform external validation/translate tools to new, independent settings. However, as reinforcement learning can already achieve high sensitivities without having to use threshold adjustment, and additionally, learn an augmented representation of a task, it may have greater ability to generalize. This was demonstrated by the results, as Q-imb most consistently achieved the highest F and G scores across different optimization thresholds and across different test sites (whereas the neural network- and XGBoost-based methods showed greater variation across thresholds and test sites). Thus, choosing a decision threshold should be carefully considered, as it directly affects F and G metrics (through the shifting of sensitivity/specificity). Future experiments could consider using bespoke thresholds adapted to each independent dataset in order to improve classification performances on each set (Yang et al., 2022b).

As technological capabilities in memory and processing continue to advance, datasets are becoming much larger, complex, and imbalanced. In this paper, we discussed one domain - healthcare - where this is imbalance is especially severe; however, the framework introduced can be adapted to many other domains. Thus, as standard methods become inadequate for coping with such extreme levels of data imbalance, novel deep learning approaches will be key to propelling forward evidence-based AI.  

\section*{Contributions}
JY conceived and designed the study, with input from AAS, ASL, and DAC. JY wrote the code, performed the main analyses, and wrote the manuscript. RE and OOD helped verify the methodology. JY \& AAS preprocessed and verified the COVID-19 datasets. AAS advised the features and categories to be used for the COVID-19 task. JY \& RE preprocessed the eICU dataset. AAS and ASL advised the features and categories to be used for the ICU discharge prediction and diagnosis tasks. All authors revised the manuscript.

\section*{Acknowledgements}
We express our sincere thanks to all patients and staff across the four participating NHS trusts; Oxford University Hospitals NHS Foundation Trust, University Hospitals Birmingham NHS Trust, Bedfordshire Hospitals NHS Foundations Trust, and Portsmouth Hospitals University NHS Trust. We additionally express our gratitude to Jingyi Wang \& Dr Jolene Atia at University Hospitals Birmingham NHS Foundation trust, Phillip Dickson at Bedfordshire Hospitals, and Paul Meredith at Portsmouth Hospitals University NHS Trust for assistance with data extraction.

\section*{Funding}
This work was supported by the Wellcome Trust/University of Oxford Medical \& Life Sciences Translational Fund (Award: 0009350) and the Oxford National Institute of Research (NIHR) Biomedical Research Campus (BRC). The funders of the study had no role in study design, data collection, data analysis, data interpretation, or writing of the manuscript. JY is a Marie Sklodowska-Curie Fellow, under the European Union’s Horizon 2020 research and innovation programme (Grant agreement: 955681, "MOIRA"). OOD is supported by the EPSRC Center for Doctoral Training in Health Data Science (EP/S02428X/1). ASL is a Rhodes Scholar and is funded by the Rhodes Trust. AAS is an NIHR Academic Clinical Fellow (Award: ACF-2020-13-015). The views expressed are those of the authors and not necessarily those of the NHS, NIHR, EU Commission, Rhodes Trust, or the Wellcome Trust.

\section*{Ethics}
United Kingdom National Health Service (NHS) approval via the national oversight/regulatory body, the Health Research Authority (HRA), has been granted for development and validation of artificial intelligence models to detect Covid-19 (CURIAL; NHS HRA IRAS ID: 281832).

The eICU Collaborative Research Database (eICU-CRD) is a publicly-available, anonymized database with pre-existing institutional review board (IRB) approval; thus, no further approval was required.

\section*{Declarations and Competing Interests}
DAC reports personal fees from Oxford University Innovation, personal fees from BioBeats, personal fees from Sensyne Health, outside the submitted work. No other authors report any conflicts of interest.

\section*{Data and Code Availability}
Data from OUH studied here are available from the Infections in Oxfordshire Research Database, subject to an application meeting the ethical and governance requirements of the Database. Data from UHB, PUH and BH are available on reasonable request to the respective trusts, subject to HRA requirements. The eICU Collaborative Research Database is available online. Code and supplementary information for this paper are available online alongside publication.

\newpage

\section*{References}

{
\small

[1] Haixiang, G., Yijing, L., Shang, J., Mingyun, G., Yuanyue, H. \& Bing, G. (2017). Learning from class-imbalanced data: Review of methods and applications. Expert systems with applications, 73, 220-239.

[2] Kaur, H., Pannu, H. S., \& Malhi, A. K. (2019). A systematic review on imbalanced data challenges in machine learning: Applications and solutions. ACM Computing Surveys (CSUR), 52(4), 1-36.

[3] Ganganwar, V. (2012). An overview of classification algorithms for imbalanced datasets. International Journal of Emerging Technology and Advanced Engineering, 2(4), 42-47.

[4] Zong, W., Huang, G. B., \& Chen, Y. (2013). Weighted extreme learning machine for imbalance learning. Neurocomputing, 101, 229-242.

[5] Denil, M., \& Trappenberg, T. (2010, May). Overlap versus imbalance. In Canadian conference on artificial intelligence (pp. 220-231). Springer, Berlin, Heidelberg.

[6] Tyagi, S., \& Mittal, S. (2020). Sampling approaches for imbalanced data classification problem in machine learning. In Proceedings of ICRIC 2019 (pp. 209-221). Springer, Cham.

[7] He, H., \& Ma, Y. (Eds.). (2013). Imbalanced learning: foundations, algorithms, and applications.

[8] Fernández, A., García, S., Galar, M., Prati, R. C., Krawczyk, B., \& Herrera, F. (2018). Learning from imbalanced data sets (Vol. 10, pp. 978-3). Berlin: Springer.

[9] Chawla, N. V., Bowyer, K. W., Hall, L. O., \& Kegelmeyer, W. P. (2002). SMOTE: synthetic minority over-sampling technique. Journal of artificial intelligence research, 16, 321-357.

[10] Chen, R. J., Lu, M. Y., Chen, T. Y., Williamson, D. F., \& Mahmood, F. (2021). Synthetic data in machine learning for medicine and healthcare. Nature Biomedical Engineering, 5(6), 493-497.

[11] Lin, E., Chen, Q., \& Qi, X. (2020). Deep reinforcement learning for imbalanced classification. Applied Intelligence, 50(8), 2488-2502.

[12] Yang, J., Soltan, A. A., Yang, Y., \& Clifton, D. A. (2022). Algorithmic Fairness and Bias Mitigation for Clinical Machine Learning: Insights from Rapid COVID-19 Diagnosis by Adversarial Learning. medRxiv.

[13] Wang, Z., Schaul, T., Hessel, M., Hasselt, H., Lanctot, M., \& Freitas, N. (2016, June). Dueling network architectures for deep reinforcement learning. In International conference on machine learning (pp. 1995-2003). PMLR.

[14] Thrun, S., \& Schwartz, A. (1993, December). Issues in using function approximation for reinforcement learning. In Proceedings of the 1993 Connectionist Models Summer School Hillsdale, NJ. Lawrence Erlbaum (Vol. 6).

[15] Van Hasselt, H., Guez, A., \& Silver, D. (2016, March). Deep reinforcement learning with double q-learning. In Proceedings of the AAAI conference on artificial intelligence (Vol. 30, No. 1).

[16] Sui, Z., Pu, Z., Yi, J., \& Tan, X. (2018, July). Path planning of multiagent constrained formation through deep reinforcement learning. In 2018 International Joint Conference on Neural Networks (IJCNN) (pp. 1-8). IEEE.

[17] Gu, Q., Zhu, L., \& Cai, Z. (2009, October). Evaluation measures of the classification performance of imbalanced data sets. In International symposium on intelligence computation and applications (pp. 461-471). Springer, Berlin, Heidelberg.

[18] Pollard, T. J., Johnson, A. E., Raffa, J. D., Celi, L. A., Mark, R. G., \& Badawi, O. (2018). The eICU Collaborative Research Database, a freely available multi-center database for critical care research. Scientific data, 5(1), 1-13.

[19] Soltan, A. A., Yang, J., Pattanshetty, R., Novak, A., Rohanian, O., Beer, S., ... \& Clifton, D. A. (2022). Real-world evaluation of rapid and laboratory-free COVID-19 triage for emergency care: external validation and pilot deployment of artificial intelligence driven screening. The Lancet Digital Health.

[20] Goudouris, E. S. (2021). Laboratory diagnosis of COVID-19. Jornal de pediatria, 97, 7-12.

[21] Yang, J., Soltan, A. A., \& Clifton, D. A. (2022). Machine Learning Generalizability Across Healthcare Settings: Insights from multi-site COVID-19 screening. medRxiv.

}

\newpage

\appendix

\section{Software Packages and Implementation}

Models were implemented using Python (v3.6.9). Scikit Learn (v0.24.1) was used for standardization, median imputation, and calculating performance metrics. Imbalanced Learn (v0.7.0) was used to implement SMOTE. Performance metrics were calculated using Scikit Learn and manually programmed. XGBoost baseline models were implemented using the XGBoost library (v1.3.3). Neural network baseline models were implemented using Keras (v2.6.0). Reinforcement learning was set up using Tensorflow (v2.6.2). All models were run using an Intel Xeon E-2146G Processor (CPU: 6 cores, 4.50 GHz max frequency). 

\section{Reinforcement Learning for Classification}
Reinforcement learning (RL) has been linked to many real-world AI applications, with some of its most well-known successes stemming from game play and control challenges (e.g. AlphaGo [Silver et al., 2017], StarCraft [Vinyals et al., 2019], Atari games [Mnih et al., 2013], etc.). However, the core elements of RL have been shown to be successful on a wider range of tasks, including those that, on the surface, do not appear to have a particular “agent” interacting with an “environment” (which is typically regarded as the standard RL set-up [Sutton \& Barto, 2018; Li, 2017]). Such problems include classification tasks, which have commonly been addressed using standard supervised learning algorithms, where an input through a model to predict a class label. RL, instead, uses an agent to interact with the input to determine which class it belongs to, and then receives an immediate reward from its environment based on that prediction.  A positive reward is given to the agent when a label is correctly predicted, and a negative one is given otherwise. This feedback helps the agent learn the optimal "behavior" for classifying samples correctly, such that it accumulates the maximum rewards. This learned behavior is an augmented representation of the task, making it possible to learn beyond the immediate information encoded in the input (Wiering et al., 2011). To do this, an agent performs actions that set memory cells, which then can be used by the agent, together with the original input, to select actions and classify samples (Wiering et al., 2011; Lin et al., 2019).     
 
Recently, a branch of RL known as Q-learning, has been found to be successful for classification tasks. Ling et al. (2017) developed a reinforcement leaning framework for patient diagnosis using clinical free text data. Here, a deep Q-network (DQN) was used, and preliminary results showed improvement over non-RL baselines. DQN was also used for classification tasks in Martinez et al. (2018), where authors evaluated its efficacy for early classification tasks using timeseries data. Similarly, they found that RL could achieve effective performance over benchmarks.

\section{Model Architectures}

\subsection{Baseline Model Architectures}
\textbf{Neural Network}: The rectified linear unit (ReLU) activation function was used for the hidden layers and the sigmoid activation function was used in the output layer. For updating model weights, the Adaptive Moment Estimation (Adam) optimizer was used during training. 

\textbf{XGBoost}: XGBoost (Chen \& Guestrin, 2016) is a popular ensemble model that has achieved state-of-the-art results on many machine learning challenges. Ensemble methods combine the predictions of multiple models, such that the generalization error is improves (i.e., contribution of individual error from any individual model is lessened). XGBoost in particular, utilizes a boosting technique, where trees are sequentially added and fit to correct for the prediction errors made by previous models. Default settings were used in all experiments.

\subsection{Reinforcement Learning Architecture}

We used a fully-connected neural network with one hidden layer, alongside the rectified linear unit (ReLU) activation function and dropout. The raw output was then fed into two, separate, advantage ($A$) and value ($V$) streams. $A$ and $V$, and additionally, a softmax activation layer (applied to $A$), are then combined to produce the final output, $Q$. For updating model weights, the Adaptive Moment Estimation (Adam) optimizer was used during training. We set the exploration probability, $\epsilon$, to be linearly attenuated from 1 to 0.01 over the entire training process. Each training period consists of 120,000 steps (i.e. iterations of updating parameters $\theta$).
\FloatBarrier
\begin{figure}[ht]
\centering
\includegraphics[scale=0.4]{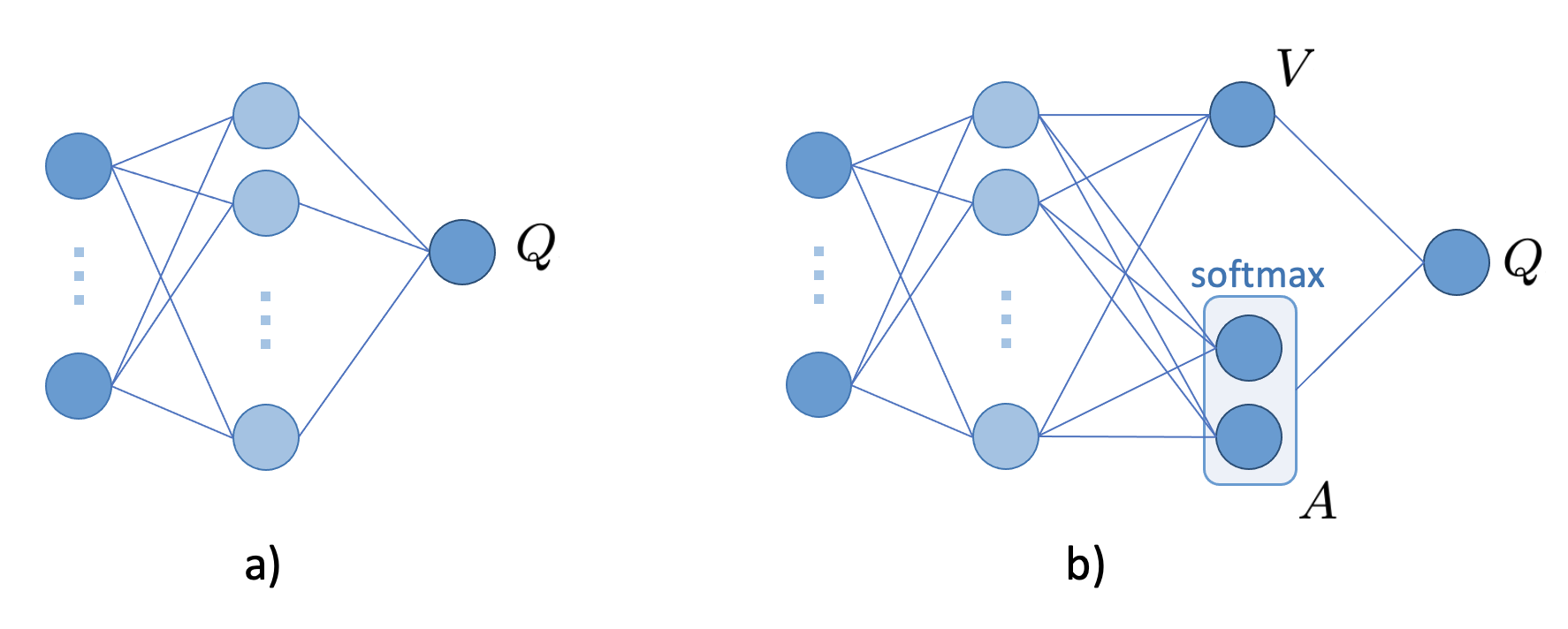}
\caption{A typical single-stream Q-network is shown in a). A dueling architecture, with two streams to independently estimate the state-values (scalar) and advantages (vector) for each each action is shown in b) (this implements equation 9).}
%\label{fig:dueling}
\end{figure}
\FloatBarrier

\section{Q-imb Training Procedure}

In each episode, the agent employs an $\epsilon$-greedy behavior policy to select an action, which randomly selects an action with probability $\epsilon$, or an action following the optimal Q-function, $arg\max_aQ^*(s_t, a_t)$ with probability $1-\epsilon$. A subsequent reward is then given from the environment through the process described in Algorithm 1. The overall Q-network is trained according to the DDQN process described in Algorithm 2. The final, optimized Q-network is considered to be the trained classifier.  

\SetKwInput{KwInput}{Input}  % Set the Input
\SetKwInput{KwOutput}{Output}

\begin{algorithm}[!ht]
\DontPrintSemicolon
  %\KwInput{Your Input}
  %\KwOutput{Your output}
  %\KwData{Testing set $x$}
  % Set Function Names
  \SetKwFunction{FMain}{Reward}
% Write Function with word ``Function''
  \SetKwProg{Fn}{Function}{:}{}
  State $D_1$ is the minority class set in the training data.\;
  \Fn{\FMain{$a_t$, $l_t$}}{
    Initialize $term_t = False$ \;
    \If{$a_t = l_t$}
    {
        Set $r_t = \lambda$
    }
    \Else
    {
        Set $r_t = -\lambda$\;
        \If{$l_t \epsilon D_1$}
        {
        Set $term = True$
        }
    }
    \KwRet $r_t$, $term$\;
  }
\caption{Environment Reward Procedure}
\end{algorithm}

\begin{algorithm}[!ht]
\DontPrintSemicolon
  %\KwInput{Your Input}
  %\KwOutput{Your output}
  %\KwData{Testing set $x$}
  % Set Function Names
  \SetKwFunction{FMain}{Main}
% Write Function with word ``Function''
  \SetKwProg{Fn}{Function}{:}{}
  Initialize memory, M\;
  \Fn{\FMain{$a_t$, $l_t$}}{
    Initialize $term_t = False$ \;
    \For{$episode \epsilon \{1, 2, ..., K\}$}
    {
        Shuffle training data, D\;
        Initialize state $s_1 = x_1$\;
        \For{$t \epsilon \{1, 2, ..., T\}$}
        {
        Choose action using $\epsilon$-greedy policy:\;
        $a_t = \pi_\theta(s_t)$\;
        $r_t$, $term_t$ = Reward($a_t$, $l_t$)\;
        Set $s_{t+1} = x_{t+1}$\;
        Store ($s_t, a_t, r_t, s_{t+1},term_t$) to M\;
        Randomly sample ($s_j, a_j, r_j, s_{j+1},term_j$) from M\;
        \If{$term_j = True$}
        {
        Set $y_j=r_j$ 
        }
        \Else
        {
        Set $y_j= r_j+\gamma Q(s_{j+1}, arg\max_a Q(s_{j+1},a_{j+1};\theta); \theta')$
        }
        Perform gradient descent on $L(\theta)$ w.r.t. $\theta$\;
        Set $\theta' = \theta$
        }
        \If{$term_t = True$}
        {
        Break
        }
    }
  }
\caption{DDQN Training Procedure}
\end{algorithm}

\FloatBarrier

\section{Evaluation Metrics}

As the purpose of our model is to effectively train models on imbalanced data, we use F-measure and G-mean metrics to evaluate the balanced classification performance of each model. These are known as the harmonic and geometric means, respectively. The harmonic mean of two values is usually closer to the smallest one; thus, a higher F value implies that sensitivity and precision are simultaneously high (Gu et al., 2009). Similarly, the geometric mean evaluates sensitivity and specificity, ensuring that both the true positive and true negative rates are high (Gu et al., 2009).

As used in Lin et al. (2019), we calculate F-measure and G-mean as follows:

\begin{equation}
\begin{aligned}
    F = &\sqrt{Sensitivity * Precision} \\
    = & \sqrt{\frac{TP}{TP+FN} * \frac{TP}{TP+FP}}
\end{aligned}
\end{equation}

\begin{equation}
\begin{aligned}
    G = &\sqrt{Sensitivity * Specificity} \\
    = &\sqrt{\frac{TP}{TP+FN} * \frac{TN}{TN+FP}}
\end{aligned}
\end{equation}

where TP is the number of true positives; FP is the number of false positives; TN is the number of true negatives; and FN is the number of false negatives.

\newpage

\section{COVID-19 Data and Preprocessing}

\subsection{Ethics statement}
United Kingdom National Health Service (NHS) approval via the national oversight/regulatory body, the Health Research Authority (HRA), has been granted for this work (IRAS ID: 281832).

\subsection{Data Inclusion and Exclusion}

\textbf{Oxford University Hospitals NHS Foundation Trust (OUH):}
We included all patients attending acute and emergency care settings at OUH who received routine blood tests on arrival, considering presentations before December 1, 2019, and thus before the pandemic, as the COVID-19-negative (control) cohort. We considered presentations during the ‘first wave’ of the UK COVID-19 pandemic (December 1, 2019 to June 30, 2020) with PCR confirmed SARS-CoV-2 infection as the COVID-19-positive (cases) cohort. We excluded patients who opted out of electronic health record (EHR) research and those who did not receive laboratory blood tests or were younger than 18 years of age. Due to incomplete penetrance of testing during the first wave of the pandemic, and imperfect sensitivity of the PCR test, there is uncertainty in the viral status of patients presenting during the pandemic who were untested or tested negative. We therefore selected a pre-pandemic control cohort during training to ensure absence of disease in patients labelled as COVID-19-negative. Clinical features extracted for each presentation included first-performed blood tests, blood gases, vital signs measurements and PCR testing for SARS-CoV-2 (Abbott Architect [Abbott, Maidenhead, UK], TaqPath [Thermo Fisher Scientific, Massachusetts, USA] and Public Health England-designed RNA-dependent RNA polymerase assays). 

\textbf{Portsmouth Hospitals NHS Foundation Trust (PUH):}
PUH considered all patients admitted to the Queen Alexandria Hospital, serving a population of 675,000 and offering tertiary referral services to the surrounding region, between March 1, 2020 and February 28, 2021. Confirmatory COVID-19 testing was by laboratory SARS-CoV2 RT-PCR assay, considering any positive PCR result within 48hrs of admission as a true positive. 

\textbf{University Hospitals Birmingham NHS Foundation Trust (UHB):}
UHB considered all patients admitted to The Queen Elizabeth Hospital, Birmingham, between December 01, 2019 and October 29, 2020. The Queen Elizabeth Hospital is a large tertiary referral unit within the UHB group which provides healthcare services for a population of 2.2 million across the West Midlands. Confirmatory COVID-19 testing was performed by laboratory SARS-CoV-2 RT-PCR assay. 

\textbf{Bedfordshire NHS Foundation Trust (BH):}
BH considered all patients admitted to Bedford Hospital between January 1, 2021 and March 31, 2021. BH provides healthcare services for a population of around 620,000 in Bedfordshire. Confirmatory COVID-19 testing was performed on the day of admission by point-of-care PCR based nucleic acid testing [SAMBA-II \& Panther Fusion System, Diagnostics in the Real World, UK, and Hologic, USA].

\FloatBarrier

\begin{table}[ht]
\caption{Summary population characteristics for OUH training cohorts, prospective validation cohort of patients attending OUH, independent validation cohorts of patients admitted to three independent NHS Trusts. *indicates merging for statistical disclosure control.}
\resizebox{\textwidth}{!}{
\begin{tabular}{@{}lllllll@{}}
\toprule
& \multicolumn{2}{l}{OUH (pre-pandemic \& wave 1 cases, to 30/06/2020)} & OUH                   & PUH                   & UHB                   & BH                    \\ \midrule
Cohort                     & Pre-pandemic cohort              & COVID-19-cases cohort              & 01/10/2020-06/03/2021 & 01/03/2020-28/02/2021 & 01/12/2019-29/10/2020 & 01/01/2021-31/03/2021 \\
\textbf{n, patients}       & 114,957                          & 701                                & 22,857                & 37,896                & 10,293                & 1177                  \\
\textbf{n, COVID positive} & 0                                & 701                                & 2,012 (8.80\%)        & 2,005 (5.29\%)        & 439 (4.27\%)          & 144 (12.2\%)          \\
\textbf{Sex:}              &                                  &                                    &                       &                       &                       &                       \\
- Male (\%)                & 53370 (46.43)                    & 376 (53.64)                        & 11409 (49.91)         & 20839 (54.99)         & 4831 (46.93)          & 627 (53.27)           \\
- Female (\%)              & 61587 (53.57)                    & 325 (46.36)                        & 11448 (50.09)         & 17054 (45.0)          & 5462 (53.07)          & 549 (46.64)           \\
Age, yr (IQR)              & 60 (38-76)                       & 72 (55-82)                         & 67 (49-80)            & 69 (48-82)            & 63 (42-79)            & 68.0 (48-82)          \\
\textbf{Ethnicity:}        &                                  &                                    &                       &                       &                       &                       \\
-White (\%)                & 93921 (81.7)                     & 480 (68.47)                        & 17387 (76.07)         & 28704 (75.74)         & 6848 (66.53)          & 1024 (87.0)           \\
-Not Stated (\%)           & 13602 (11.83)                    & 128 (18.26)                        & 4127 (18.06)          & 8389 (22.14)          & 1061 (10.31)          & $\le$10                  \\
-South Asian (\%)          & 2754 (2.4)                       & 22 (3.14)                          & 441 (1.93)            & 170 (0.45)            & 1357 (13.18)          & 71 (6.03)             \\
-Chinese (\%)              & 284 (0.25)                       & *                                  & 51 (0.22)             & 42 (0.11)             & 41 (0.4)              & $\le$10                  \\
-Black (\%)                & 1418 (1.23)                      & 25 (3.57)                          & 279 (1.22)            & 187 (0.49)            & 484 (4.7)             & 36 (3.06)             \\
-Other (\%)                & 1840 (1.6)                       & 34 (4.85)*                         & 410 (1.79)            & 269 (0.71)            & 333 (3.24)            & 29 (2.46)             \\
-Mixed (\%)                & 1138 (0.99)                      & 12 (1.71)                          & 162 (0.71)            & 135 (0.36)            & 169 (1.64)            & 13 (1.1)              \\ \bottomrule
\end{tabular}%
}
\end{table}

\begin{table}[ht]
\begin{center}
\caption{Clinical predictors considered.}
%\label{table}
%\setlength{\tabcolsep}{2pt}
\scalebox{0.9}{
\begin{tabular}{p{0.43\textwidth}p{0.5\textwidth}}\toprule
%\begin{tabular}{|l|l|}

\multicolumn{1}{c}{\textbf{Category}} & \multicolumn{1}{c}{\textbf{Features}}             \\ \midrule
Vital Signs &
  Heart rate, respiratory rate, systolic blood pressure, diastolic blood pressure, temperature \\ 
Full Blood Count &
  Haemoglobin, haematocrit, mean cell volume, white cell count, neutrophil count, lymphocyte count, monocyte count, eosinophil count, basophil count, platelets \\ 
Liver Function Tests \& C-reactive protein        & Albumin, alkaline phosphatase, alanine aminotransferase, bilirubin, C-reactive protein \\ 
Urea \& Electrolytes            & Sodium, potassium, creatinine, urea, estimated glomerular filtration rate         \\ \bottomrule
\end{tabular}
}
\end{center}
%\label{tab1}
\end{table}

\FloatBarrier

\subsection{Preprocessing}
We used electronic health record (EHR) data with linked, deidentified demographic information for all patients presenting to emergency departments. To better compare our results to previously published studies using the same datasets (Soltan et al., 2022, Yang et al., 2022a, Yang et al., 2022b), we used the same focused subset of routinely collected clinical features (including blood tests and vital signs) and patient cohorts.

The OUH training set consisted of COVID-free cases prior to the outbreak, so we matched every COVID-positive case to twenty COVID-free presentations based on age, representing a simulated prevalence of 5\%. Consistent with previous studies, we also used population median imputation to replace any missing values. We then standardized all features in our data to have a mean of 0 and a standard deviation of 1. 

A training set was used for model development, hyperparameter selection, and training; a validation set was used for threshold-adjustment; and after successful development and training, held-out test sets were then used to evaluate the performance of the final model.

\begin{table}[ht]
\centering
\caption{Previously published COVID-19 status prediction results. using same datasets and patient cohorts. Sensitivity, specificity, and AUROC shown, alongside 95\% confidence intervals, unless otherwise specified.}
%\resizebox{\textwidth}{!}{%
\begin{tabular}{@{}llll@{}}
\toprule
Test Set       & Sensitivity              & Specificity              & AUROC                    \\ \midrule
\multicolumn{4}{l}{\textbf{Soltan et al., 2022.}}                                               \\
\multicolumn{4}{l}{\textit{Method: XGBoost + SMOTE + Threshold Adjustment (0.9)}}               \\
OUH            & 0.857 (SD 0.009)         & 0.686 (SD 0.022)         & 0.878 (SD 0.001)         \\
PUH            & 0.841 (0.825-0.857)      & 0.713 (0.709 -0.718)     & 0.872 (0.863 -0.882)     \\
UHB            & 0.788 (0.748-0.824)      & 0.747 (0.738 -0.755)     & 0.858 (0.838 -0.878)     \\
BH             & 0.743 (0.666-0.807)      & 0.848 (0.825 0. 869)     & 0.881 (0.851- 0.912)     \\ \midrule
\multicolumn{4}{l}{\textbf{Yang et al., 2022.}}                                                 \\
\multicolumn{4}{l}{\textit{Method: Neural Network + SMOTE + ENN + Threshold Adjustment (0.85)}} \\
OUH            & 0.844 (0.828-0.860)      & 0.710 (0.704-0.717)      & 0.777 (0.765-0.789)      \\
PUH            & 0.857 (0.842-0.873)      & 0.672 (0.667-0.677)      & 0.765 (0.752-0.777)      \\
UHB            & 0.847 (0.814-0.881)      & 0.716 (0.708-0.725)      & 0.782 (0.756-0.808)      \\
BH             & 0.847 (0.789-0.906)      & 0.822 (0.799-0.845)      & 0.835 (0.793-0.876)      \\ \midrule
\multicolumn{4}{l}{\textbf{Yang et al., 2022.}}                                               \\
\multicolumn{4}{l}{\textit{Method: Neural Network + Threshold Adjustment (0.85)}}               \\
OUH            & 0.762 (0.744-0.781)      & 0.844 (0.839-0.849)      & 0.878 (0.868-0.888)      \\
PUH            & 0.633 (0.585-0.681)      & 0.903 (0.897-0.910)      & 0.861 (0.837-0.885)      \\
UHB            & 0.714 (0.621-0.807)      & 0.854 (0.839-0.870)      & 0.878 (0.832-0.924)      \\
BH             & 0.724 (0.561-0.887)      & 0.908 (0.869-0.948)      & 0.880 (0.798-0.963)      \\ \bottomrule
\end{tabular}%
%}
\end{table}

\section{Multiclass Patient Diagnosis Data and Preprocessing}

\subsection{Ethics statement}
The eICU Collaborative Research Database (eICU-CRD) (Pollard et al., 2018) is a publicly-available, anonymized database with pre-existing institutional review board (IRB) approval; thus, no further approval was required.

\subsection{Data Inclusion and Exclusion}
In terms of clinical applications of AI, patient diagnosis as been a popular problem to address (Sheikhalishahi et al., 2021; Lipton et al., 2015; Razavian et al., 2016), as it can directly influence clinical decision-making, resource allocation, and healthcare costs.

Here, the task was to predict which acute condition might be developed by a patient during the course of an ICU stay, as defined through ICD-9 codes. A similar task that included both acute and chronic conditions was previously investigated using the eICU-CRD dataset by grouping 767 ICD-9 codes into 25 overarching diagnoses, and then predicting these using a BiLSTM model (Sheikhalishahi et al., 2021). Using similar inclusion and exclusion criteria, we selected adult patients (age > 18) with a minimum of 15 ICU records, and grouped these records into 1 hour windows. Our clinical team reviewed the list of 25 diagnoses, removed 13 diagnoses considered chronic, non-acute, or poorly defined, and grouped the remaining 12 diagnoses into their relevant system and clinical specialties. This resulted in five labels: acute cardiovascular event, acute respiratory event, acute gastrointestional event, acute systemic event, and acute renal event. This grouping was selected to reflect clinic reality, where an emergency physician might consult with a system specialist to rule out a severe condition before admission to ICU, and to account for the relatedness of diagnoses within a system. For example, pneumonia is a leading cause of respiratory failure, and combining both diagnoses into a single "acute respiratory event" category reflects the systemic nature of the disease. We removed any samples that did not have a differentiable ICD9 code, or did not belong to any of the curated groups, resulting in 24,102 samples for training and testing.

\begin{table*}[ht]
\begin{center}
\caption{Summary population characteristics for eICU-CRD cohort.}
%\label{table}
\begin{tabular}{ll} \toprule

\textbf{Characteristic} & \textbf{eICU-CRD Cohort} \\ \midrule
\begin{tabular}[c]{@{}l@{}}Sex:\\ Female (\%)\end{tabular} & 10842 (45.0) \\ 
Male (\%) & 13260 (55.0) \\ 
Age (IQR) & 65 (54-76) \\ 
\begin{tabular}[c]{@{}l@{}}Ethnicity:\\ Unknown (\%)\end{tabular} & 1462 (6.1) \\ 
Asian (\%) & 479 (2.0) \\ 
African American (\%) & 2557 (10.6) \\ 
Caucasian (\%) & 18440 (76.5) \\ 
Hispanic (\%) & 1002 (4.2) \\ 
Native American (\%) & 162 (0.67) \\ \bottomrule
\end{tabular}
\end{center}
\label{tab2}
\end{table*}

\begin{table}[ht]
\begin{center}
\caption{Acute event groups and respective prevalences.}
%\label{table}
%\setlength{\tabcolsep}{2pt}
\begin{tabular}{cp{150pt}c} \toprule

\textbf{Label} & \textbf{Events} & \textbf{Prevalence} \\ \midrule
Acute cardiovascular event & Acute myocardial infarction,   acute cerebrovascular disease & 0.288 \\ 
Acute respiratory event & Respiratory failure, insufficiency, arrest, pneumonia, pleurisy, pneumothorax, pulmonary collapse, other upper respiratory disease, other lower respiratory disease & 0.336 \\ 
Acute gastrointestinal event & Gastrointestinal hemorrhage & 0.087 \\ 
Acute systemic event & Septicemia, shock & 0.174 \\ 
Acute renal event & Renal failure, fluid and   electrolyte disorder & 0.113 \\ \bottomrule
\end{tabular}
\end{center}
%\label{tab1}
\end{table}

\begin{table}[ht]
\begin{center}
\caption{Clinical predictors considered for predicting patient discharge status and patient diagnosis.}
\label{table}
\begin{tabular}{lp{170pt}} \toprule

\textbf{Category} & \textbf{Features} \\ \midrule
Demographic features & Gender, age, ethnicity, height,   weight \\ 
Measurements at hospital   admission & Non-invasive systolic   blood pressure, non-invasive diastolic blood pressure, non-invasive mean   arterial pressure, heart rate, Supporting oxygen used at   admission, blood oxygen saturation, Glasgow coma score, diagnosis at   admission \\ 
Measurements at ICU   admission & Glucose \\ \bottomrule
\end{tabular}
\end{center}
%\label{tab1}
\end{table}

\FloatBarrier

\subsection{Preprocessing}
Further preprocessing was performed to remove samples with any missing values, one-hot encode categorical features, and standardize all continuous features to have a mean of 0 and a standard deviation of 1. 

We used a 75:25 training and test ratio, resulting in 18,076 training and 6,026 test samples, respectively. As before, the training set was used for model development, hyperparameter selection, and training; and after successful development and training, the held-out test set was used to evaluate the performance of the final model. It should be noted that, as this is a multiclass task, standard threshold adjustment cannot be used, and thus, we did not split the data to include an additional validation set.

\section{Training, Thresholds, and Hyperparameter Values}

\subsection{COVID-19 Diagnosis}

\begin{table}[ht]
\caption{Final Hyperparameter Values Used in Reinforcement Learning, Neural Network, and XGBoost-Based Models in COVID-19 Prediction Task}
\begin{center}
%\label{table}
\begin{tabular}{ccc}\toprule

\textbf{Reinforcement Learning} & \textbf{Neural Network} & \textbf{XGBoost} \\ \midrule
\begin{tabular}[c]{@{}c@{}}Dropout = 0.3\\ Learning rate = 0.0004\\ Neurons = 100\\ Discount Factor = 0.1\\ Exploration Factor = {[}0.01,   1{]}\end{tabular} & \begin{tabular}[c]{@{}c@{}}Dropout = 0.3\\ Learning rate = 0.1\\ Neurons = 10\end{tabular} & \begin{tabular}[c]{@{}c@{}}Depth = 3\\ N estimators = 100  \\ Learning rate = 0.1\end{tabular} \\ \bottomrule
\end{tabular}
%\label{tab4}
\end{center}
\end{table}

\FloatBarrier

We implemented early stopping, which monitored validation performance, optimizing training for a sensitivity of $>$0.85 and specificity of $>$0.75. These thresholds were set to ensure that the model would be able to detect positive COVID-19 cases.

\begin{table}[ht]
\caption{Adjusted Threshold Values Used in Reinforcement Learning, Neural Network, and XGBoost Models, for COVID-19 status prediction.}
\centering
\begin{tabular}{@{}lll@{}}
\toprule
Model & \multicolumn{2}{l}{Threshold} \\ \midrule
 & 0.85 & 0.9 \\ \cmidrule(l){2-3} 
Reinforcement Learning (Q-imb) & 0.5050 & 0.4970 \\
Reinforcement Learning (DDQN) & 0.5070 & 0.5010 \\
XGBoost & 0.0060 & 0.0020 \\
XGBoost + SMOTE & 0.0120 & 0.0060 \\
XGBoost + Cost-Sensitive & 0.0120 & 0.0060 \\
NN & 0.0341 & 0.0140 \\
NN + Cost-Senstive & 0.4148 & 0.2645 \\
NN + SMOTE & 0.0862 & 0.0641 \\ \bottomrule
\end{tabular}
\end{table}

\subsection{Multiclass Patient Diagnosis}

\begin{table}[ht]
\caption{Final Hyperparameter Values Used in Reinforcement Learning, Neural Network, and XGBoost-Based Models in COVID-19 Prediction Task}
\begin{center}
%\label{table}
\begin{tabular}{ccc} \toprule
\textbf{Reinforcement Learning} & \textbf{Neural Network} & \textbf{XGBoost} \\ \midrule
\begin{tabular}[c]{@{}c@{}}Dropout = 0.3\\ Learning rate = 0.0001\\ Neurons = 3000 \\ Discount Factor = 0.1\\ Exploration Factor = {[}0.01,   1{]}\end{tabular} & \begin{tabular}[c]{@{}c@{}}Dropout = 0.3\\ Learning rate = 0.01 \\ Neurons = 200\end{tabular} & \begin{tabular}[c]{@{}c@{}}Depth = 3\\ N estimators = 100  \\ Learning rate = 0.1\end{tabular} \\ \bottomrule
\end{tabular}
%\label{tab4}
\end{center}
\end{table}

\section{Additional Results}

\subsection{DDQN and Dueling DDQN Comparison}

The dueling DDQN consistently outperforms the DDQN, across all four test sets. This can also be seen in the training curves, as the dueling DDQN appears to be able to learn a better policy.

\begin{figure}[ht]
\centering
\includegraphics[scale=0.3]{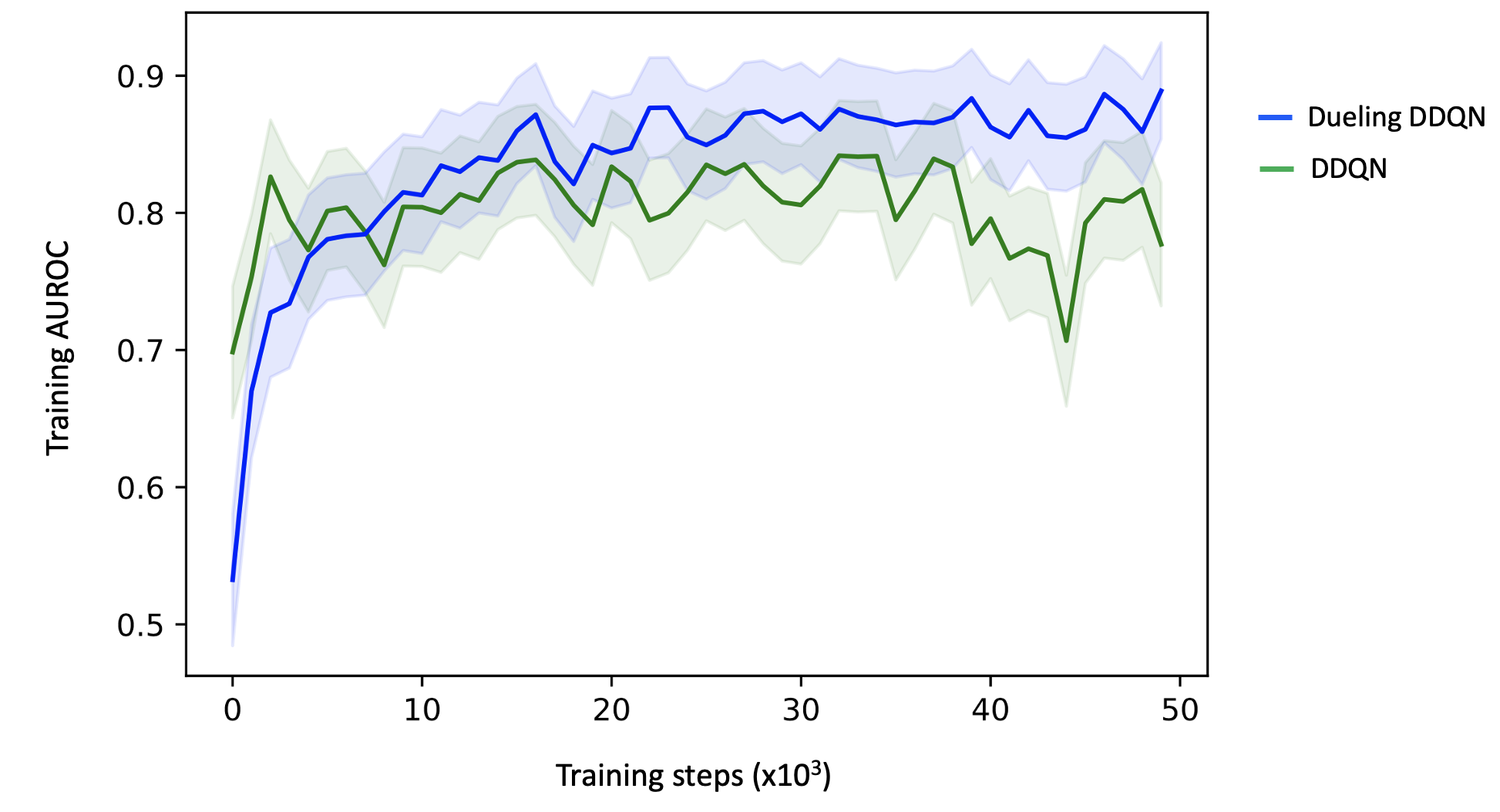}
\caption{AUROC scores during training, comparing DDQN and Dueling DDQN models. Curves are shown for the COVID-19 prediction task.}
%\label{fig:duelingcomp}
\end{figure}

\FloatBarrier

\subsection{COVID-19 Diagnosis}

\begin{table}[ht]
\caption{Performance metrics for COVID-19 prediction. Results reported as F-measure, G-mean, sensitivity, specificity, and AUROC for OUH, PUH, UHB, and BH test sets; 95\% confidence intervals (CIs) also shown. Red and blue values denote best and second best scores, respectively for F-measure and G-mean. No threshold adjustment was applied (i.e. default threshold of 0.5 used for prediction)}
\resizebox{\textwidth}{!}{%
\begin{tabular}{@{}llllll@{}}
\toprule
Model                           & F                                     & G                                     & Sensitivity         & Specificity         & AUROC               \\ \midrule
\multicolumn{6}{l}{\textbf{OUH}}                                                                                                                                                  \\
Reinforcement Learning (Q-imb) & 0.441                                 & {\color[HTML]{3166FF} \textbf{0.782}} & 0.819 (0.802-0.835) & 0.746 (0.740-0.752) & 0.861 (0.850-0.871) \\
Reinforcement Learning (DDQN)   & 0.304                                 & 0.515                                 & 0.863 (0.848-0.878) & 0.308 (0.301-0.314) & 0.758 (0.745-0.771) \\
Neural Network                  & 0.470                                 & 0.502                                 & 0.253 (0.234-0.272) & 0.996 (0.996-0.997) & 0.877 (0.867-0.886) \\
Neural Network + SMOTE          & 0.593                                 & 0.706                                 & 0.510 (0.489-0.532) & 0.978 (0.976-0.980) & 0.871 (0.861-0.881) \\
Neural Network + Cost-Sensitive & 0.507                                 & {\color[HTML]{CB0000} \textbf{0.795}} & 0.723 (0.703-0.742) & 0.874 (0.869-0.878) & 0.872 (0.862-0.882) \\
XGBoost                         & 0.577                                 & 0.623                                 & 0.391 (0.369-0.412) & 0.994 (0.992-0.995) & 0.877 (0.867-0.887) \\
XGBoost + SMOTE                 & {\color[HTML]{CB0000} \textbf{0.595}} & 0.657                                 & 0.435 (0.414-0.457) & 0.990 (0.989-0.992) & 0.876 (0.866-0.886) \\
XGBoost + Cost-Sensitive        & {\color[HTML]{3166FF} \textbf{0.584}} & 0.685                                 & 0.478 (0.456-0.499) & 0.982 (0.980-0.983) & 0.869 (0.859-0.879) \\ \midrule
\multicolumn{6}{l}{\textbf{PUH}}                                                                                                                                                  \\
Reinforcement Learning (Q-imb) & 0.316                                 & {\color[HTML]{3166FF} \textbf{0.741}} & 0.814 (0.797-0.831) & 0.674 (0.669-0.678) & 0.831 (0.819-0.842) \\
Reinforcement Learning (DDQN)   & 0.245                                 & 0.594                                 & 0.810 (0.793-0.827) & 0.435 (0.430-0.440) & 0.762 (0.750-0.774) \\
Neural Network                  & 0.47                                  & 0.601                                 & 0.367 (0.345-0.388) & 0.987 (0.985-0.988) & 0.857 (0.847-0.868) \\
Neural Network + SMOTE          & 0.452                                 & 0.735                                 & 0.574 (0.552-0.596) & 0.942 (0.940-0.944) & 0.856 (0.845-0.866) \\
Neural Network + Cost-Sensitive & 0.356                                 & {\color[HTML]{CB0000} \textbf{0.775}} & 0.767 (0.748-0.785) & 0.784 (0.780-0.789) & 0.850 (0.839-0.861) \\
XGBoost                         & {\color[HTML]{3166FF} \textbf{0.504}} & 0.639                                 & 0.414 (0.393-0.436) & 0.985 (0.984-0.987) & 0.881 (0.871-0.891) \\
XGBoost + SMOTE                 & {\color[HTML]{CB0000} \textbf{0.521}} & 0.702                                 & 0.506 (0.484-0.528) & 0.976 (0.974-0.977) & 0.881 (0.871-0.890) \\
XGBoost + Cost-Sensitive        & {\color[HTML]{CB0000} \textbf{0.521}} & 0.734                                 & 0.557 (0.535-0.578) & 0.967 (0.966-0.969) & 0.881 (0.871-0.891) \\ \midrule
\multicolumn{6}{l}{\textbf{UHB}}                                                                                                                                                  \\
Reinforcement Learning (Q-imb) & 0.315                                 & {\color[HTML]{3166FF} \textbf{0.772}} & 0.793 (0.755-0.831) & 0.753 (0.744-0.761) & 0.837 (0.814-0.861) \\
Reinforcement Learning (DDQN)   & 0.206                                 & 0.494                                 & 0.845 (0.811-0.879) & 0.289 (0.280-0.298) & 0.721 (0.694-0.749) \\
Neural Network                  & 0.351                                 & 0.444                                 & 0.198 (0.161-0.235) & 0.995 (0.993-0.996) & 0.866 (0.844-0.888) \\
Neural Network + SMOTE          & {\color[HTML]{3166FF} \textbf{0.418}} & 0.667                                 & 0.460 (0.414-0.507) & 0.966 (0.963-0.970) & 0.850 (0.828-0.873) \\
Neural Network + Cost-Sensitive & 0.364                                 & {\color[HTML]{CB0000} \textbf{0.780}} & 0.704 (0.661-0.747) & 0.864 (0.858-0.871) & 0.861 (0.839-0.883) \\
XGBoost                         & 0.417                                 & 0.548                                 & 0.303 (0.260-0.346) & 0.990 (0.988-0.992) & 0.861 (0.839-0.883) \\
XGBoost + SMOTE                 & {\color[HTML]{CB0000} \textbf{0.431}} & 0.608                                 & 0.376 (0.331-0.421) & 0.983 (0.980-0.985) & 0.853 (0.830-0.876) \\
XGBoost + Cost-Sensitive        & 0.410                                 & 0.633                                 & 0.412 (0.366-0.458) & 0.973 (0.970-0.976) & 0.851 (0.829-0.874) \\ \midrule
\multicolumn{6}{l}{\textbf{BH}}                                                                                                                                                   \\
Reinforcement Learning (Q-imb) & {\color[HTML]{3166FF} \textbf{0.582}} & {\color[HTML]{CB0000} \textbf{0.823}} & 0.799 (0.733-0.864) & 0.849 (0.827-0.871) & 0.867 (0.829-0.906) \\
Reinforcement Learning (DDQN)   & {\color[HTML]{000000} 0.364}          & 0.573                                 & 0.826 (0.765-0.888) & 0.397 (0.367-0.427) & 0.706 (0.656-0.756) \\
Neural Network                  & 0.306                                 & 0.353                                 & 0.125 (0.071-0.179) & 0.994 (0.990-0.999) & 0.885 (0.849-0.921) \\
Neural Network + SMOTE          & 0.481                                 & 0.549                                 & 0.306 (0.230-0.381) & 0.986 (0.979-0.993) & 0.882 (0.845-0.919) \\
Neural Network + Cost-Sensitive & {\color[HTML]{CB0000} \textbf{0.586}} & {\color[HTML]{3166FF} \textbf{0.759}} & 0.618 (0.539-0.697) & 0.931 (0.916-0.947) & 0.883 (0.847-0.920) \\
XGBoost                         & 0.457                                 & 0.498                                 & 0.250 (0.179-0.321) & 0.993 (0.988-0.998) & 0.894 (0.859-0.929) \\
XGBoost + SMOTE                 & 0.491                                 & 0.526                                 & 0.278 (0.205-0.351) & 0.994 (0.990-0.999) & 0.885 (0.849-0.921) \\
XGBoost + Cost-Sensitive        & 0.556                                 & 0.614                                 & 0.382 (0.303-0.461) & 0.987 (0.981-0.994) & 0.889 (0.854-0.925) \\ \bottomrule
\end{tabular}%
}
\end{table}

\FloatBarrier

\begin{table}[ht]
\caption{Performance metrics for COVID-19 prediction. Results reported as F-measure, G-mean, sensitivity, specificity, and AUROC for OUH, PUH, UHB, and BH test sets; 95\% confidence intervals (CIs) also shown. Red and blue values denote best and second best scores, respectively for F-measure and G-mean. Threshold adjustment applied to optimize models to sensitivities of 0.85.}
\resizebox{\textwidth}{!}{%
\begin{tabular}{@{}llllll@{}}
\toprule
Model                                                & F                                     & G                                     & Sensitivity         & Specificity         & AUROC               \\ \midrule
\multicolumn{6}{l}{\textbf{OUH}}                                                                                                                                                                       \\
Reinforcement Learning (Q-imb)                      & {\color[HTML]{3166FF} \textbf{0.462}} & {\color[HTML]{CB0000} \textbf{0.792}} & 0.791 (0.773-0.809) & 0.793 (0.788-0.799) & 0.861 (0.850-0.871) \\
Reinforcement Learning (DDQN)                        & 0.323                                 & 0.628                                 & 0.788 (0.770-0.806) & 0.500 (0.493-0.507) & 0.758 (0.745-0.771) \\
Neural Network                                       & 0.442                                 & 0.782                                 & 0.814 (0.797-0.831) & 0.751 (0.746-0.757) & 0.874 (0.864-0.884) \\
Neural Network + SMOTE                               & 0.412                                 & 0.756                                 & 0.837 (0.821-0.853) & 0.683 (0.676-0.689) & 0.869 (0.859-0.879) \\
Neural Network + Cost-Sensitive                      & {\color[HTML]{CB0000} \textbf{0.470}} & {\color[HTML]{CB0000} \textbf{0.792}} & 0.768 (0.750-0.787) & 0.816 (0.811-0.822) & 0.872 (0.862-0.882) \\
XGBoost                                              & 0.456                                 & 0.791                                 & 0.810 (0.793-0.827) & 0.773 (0.767-0.779) & 0.877 (0.867-0.887) \\
XGBoost + SMOTE                                      & 0.457                                 & {\color[HTML]{3166FF} \textbf{0.790}} & 0.794 (0.777-0.812) & 0.786 (0.780-0.791) & 0.876 (0.866-0.886) \\
XGBoost + Cost-Sensitive                             & 0.434                                 & 0.776                                 & 0.819 (0.802-0.836) & 0.736 (0.730-0.742) & 0.869 (0.859-0.879) \\ \midrule
\multicolumn{6}{l}{\textbf{PUH}}                                                                                                                                                                       \\
Reinforcement Learning (Q-imb)                      & 0.327                                 & 0.754                                 & 0.794 (0.776-0.812) & 0.716 (0.711-0.720) & 0.831 (0.819-0.842) \\
Reinforcement Learning (DDQN)                        & 0.262                                 & 0.661                                 & 0.756 (0.737-0.774) & 0.578 (0.573-0.584) & 0.762 (0.750-0.774) \\
Neural Network                                       & 0.331                                 & 0.760                                 & 0.825 (0.808-0.842) & 0.699 (0.695-0.704) & 0.860 (0.849-0.870) \\
Neural Network + SMOTE                               & 0.320                                 & 0.746                                 & 0.818 (0.802-0.835) & 0.681 (0.676-0.685) & 0.853 (0.842-0.863) \\
Neural Network + Cost-Sensitive                      & 0.332                                 & 0.76                                  & 0.809 (0.792-0.826) & 0.714 (0.709-0.718) & 0.850 (0.839-0.861) \\
XGBoost                                              & {\color[HTML]{3166FF} \textbf{0.363}} & {\color[HTML]{CB0000} \textbf{0.792}} & 0.836 (0.820-0.852) & 0.751 (0.746-0.755) & 0.881 (0.871-0.891) \\
XGBoost + SMOTE                                      & 0.351                                 & {\color[HTML]{3166FF} \textbf{0.781}} & 0.834 (0.818-0.851) & 0.730 (0.726-0.735) & 0.881 (0.871-0.890) \\
XGBoost + Cost-Sensitive                             & {\color[HTML]{CB0000} \textbf{0.364}} & {\color[HTML]{CB0000} \textbf{0.792}} & 0.829 (0.813-0.846) & 0.757 (0.752-0.761) & 0.881 (0.871-0.891) \\ \midrule
\multicolumn{6}{l}{\textbf{UHB}}                                                                                                                                                                       \\
Reinforcement Learning (Q-imb)                      & {\color[HTML]{3166FF} \textbf{0.327}} & 0.778                                 & 0.765 (0.726-0.805) & 0.790 (0.782-0.798) & 0.837 (0.814-0.861) \\
Reinforcement Learning (DDQN)                        & 0.214                                 & 0.598                                 & 0.761 (0.721-0.801) & 0.471 (0.461-0.480) & 0.721 (0.694-0.749) \\
Neural Network                                       & 0.325                                 & {\color[HTML]{CB0000} \textbf{0.785}} & 0.820 (0.784-0.856) & 0.752 (0.744-0.761) & 0.867 (0.846-0.889) \\
Neural Network + SMOTE                               & 0.302                                 & 0.763                                 & 0.825 (0.789-0.860) & 0.705 (0.696-0.714) & 0.848 (0.825-0.871) \\
Neural Network + Cost-Sensitive                      & {\color[HTML]{CB0000} \textbf{0.336}} & {\color[HTML]{3166FF} \textbf{0.782}} & 0.756 (0.716-0.796) & 0.808 (0.801-0.816) & 0.861 (0.839-0.883) \\
XGBoost                                              & 0.321                                 & 0.774                                 & 0.770 (0.731-0.809) & 0.779 (0.770-0.787) & 0.861 (0.839-0.883) \\
XGBoost + SMOTE                                      & 0.312                                 & 0.763                                 & 0.754 (0.714-0.794) & 0.773 (0.764-0.781) & 0.853 (0.830-0.876) \\
XGBoost + Cost-Sensitive                             & 0.308                                 & 0.764                                 & 0.774 (0.735-0.814) & 0.753 (0.745-0.762) & 0.851 (0.829-0.874) \\ \midrule
\multicolumn{6}{l}{\textbf{BH}}                                                                                                                                                                        \\
Reinforcement Learning (Q-imb) & {\color[HTML]{CB0000} \textbf{0.617}} & {\color[HTML]{CB0000} \textbf{0.837}} & 0.799 (0.733-0.864) & 0.878 (0.858-0.898) & 0.867 (0.829-0.906) \\
Reinforcement Learning (DDQN)                        & 0.349                                 & 0.625                                 & 0.660 (0.582-0.737) & 0.592 (0.562-0.622) & 0.706 (0.656-0.756) \\
Neural Network                                       & 0.589                                 & 0.820                                 & 0.778 (0.710-0.846) & 0.865 (0.845-0.886) & 0.885 (0.849-0.921) \\
Neural Network + SMOTE                               & 0.553                                 & 0.801                                 & 0.764 (0.695-0.833) & 0.840 (0.818-0.863) & 0.879 (0.842-0.916) \\
Neural Network + Cost-Sensitive                      & {\color[HTML]{3166FF} \textbf{0.614}} & 0.815                                 & 0.736 (0.664-0.808) & 0.902 (0.884-0.920) & 0.883 (0.847-0.920) \\
XGBoost                                              & 0.575                                 & {\color[HTML]{3166FF} \textbf{0.822}} & 0.806 (0.741-0.870) & 0.838 (0.816-0.861) & 0.894 (0.859-0.929) \\
XGBoost + SMOTE                                      & 0.556                                 & 0.797                                 & 0.743 (0.672-0.814) & 0.855 (0.833-0.876) & 0.885 (0.849-0.921) \\
XGBoost + Cost-Sensitive                             & 0.551                                 & 0.811                                 & 0.812 (0.749-0.876) & 0.810 (0.786-0.834) & 0.889 (0.854-0.925) \\ \bottomrule
\end{tabular}%
}
\end{table}

\FloatBarrier

\subsection{Multiclass Patient Diagnosis}

\begin{table}[ht]
\caption{Individual sensitivities per acute event class (alongside 95\% CIs), calculated using a “one-vs-all” method.}
\centering
\resizebox{\textwidth}{!}{%
\begin{tabular}{@{}cccccc@{}}
\toprule
Label & Cardiovascular & Respiratory & Gastrointestinal & Systemic & Renal \\ \midrule
Reinforcement Learning (Q-imb) & 0.853 (0.837-0.869) & 0.677 (0.657-0.697) & 0.897 (0.871-0.923) & 0.767 (0.741-0.793) & 0.545 (0.508-0.582) \\
Reinforcement Learning (DDQN) & 0.769 (0.749-0.789) & 0.303 (0.283-0.323) & 0.907 (0.882-0.932) & 0.458 (0.428-0.488) & 0.567 (0.530-0.604 \\
Neural Network & 0.900 (0.886-0.914) & 0.754 (0.735-0.773) & 0.895 (0.869-0.921) & 0.606 (0.576-0.636) & 0.419 (0.382-0.456) \\
Neural Network + SMOTE & 0.901 (0.887-0.915) & 0.783 (0.765-0.801) & 0.893 (0.866-0.920) & 0.592 (0.562-0.622) & 0.400 (0.363-0.437) \\
Neural Network + Cost-Sensitive & 0.926 (0.914-0.938) & 0.524 (0.502-0.546) & 0.905 (0.880-0.930) & 0.742 (0.715-0.769) & 0.465 (0.428-0.502) \\
XGBoost & 0.864 (0.848-0.880) & 0.849 (0.833-0.865) & 0.897 (0.871-0.923) & 0.633 (0.604-0.662) & 0.423 (0.386-0.460) \\
XGBoost + SMOTE & 0.863 (0.847-0.879) & 0.853 (0.838-0.868) & 0.903 (0.877-0.929) & 0.626 (0.597-0.655) & 0.422 (0.385-0.459) \\
XGBoost + Cost-Sensitive & 0.861 (0.845-0.877) & 0.664 (0.643-0.685) & 0.907 (0.882-0.932) & 0.774 (0.749-0.799) & 0.515 (0.478-0.552) \\ \bottomrule
\end{tabular}%
}
\end{table}

\FloatBarrier

\begin{table}[ht]
\caption{Individual G values per acute event class, calculated using a “one-vs-all” method.}
\centering
\resizebox{\textwidth}{!}{%
\begin{tabular}{@{}cccccc@{}}
\toprule
Label & Cardiovascular & Respiratory & Gastrointestinal & Systemic & Renal \\ \midrule
Reinforcement Learning (Q-imb) & 0.906 & 0.780 & 0.938 & 0.830 & 0.714 \\
Reinforcement Learning (DDQN) & 0.799 & 0.522 & 0.931 & 0.651 & 0.681 \\
Neural Network & 0.889 & 0.803 & 0.941 & 0.754 & 0.643 \\
Neural Network + SMOTE & 0.891 & 0.812 & 0.940 & 0.748 & 0.630 \\
Neural Network + Cost-Sensitive & 0.873 & 0.703 & 0.944 & 0.817 & 0.669 \\
XGBoost & 0.912 & 0.823 & 0.942 & 0.770 & 0.647 \\
XGBoost + SMOTE & 0.912 & 0.824 & 0.945 & 0.767 & 0.646 \\
XGBoost + Cost-Sensitive & 0.911 & 0.769 & 0.947 & 0.830 & 0.694 \\ \bottomrule
\end{tabular}%
}
\end{table}

\FloatBarrier

\subsection{Additional Discussion}

When threshold adjustment is not applied, we found that reinforcement learning was the only method that achieved clinically-effective sensitivities for COVID-19 prediction. We found that both baseline methods (neural network and XGBoost without any imbalanced learning strategy applied), achieved poor sensitivities for COVID-19 prediction, with the XGBoost baseline achieving slightly better results. This was also found to be the case in the multiclass diagnosis task, as the baseline methods tended to have poorer performance on minority classes, with the XGBoost performing slightly better. This is expected, as most standard supervised learning models assume that classes are equally distributed (Ganganwar, 2012; Zong et al., 2013); and thus, skewed distributions can negatively affect the minority class (Haixiang et al., 2017; Kaur et al., 2019). The XGBoost baseline model may have achieved slightly better sensitivity, as it is an ensemble method, which inherently combines the predictions of multiple models, improving the generalization error. However, sensitivity is still low, as ensemble methods still require a base classifier, which is usually a conventional machine learning model that typically isn’t suitable for imbalanced data (Haixiang et al., 2017).

When SMOTE was added to training, both the neural network and the XGBoost model only improved slightly in detecting COVID-19 cases (mean sensitivities still $<$0.5). This may be because the added synthetic examples helped create larger (and thus, less specific) decision regions, making generalization easier. However, the improvement may have been minimal because new examples could have been generated from overlapping regions. This is especially relevant to health-related tasks, as clinical data is heterogeneous, and understanding how social, behavioral, and genetic factors collectively and independently impact outcomes is difficult (Yang et al., 2021a). Thus, it can be hard to confidently augment data, resulting in noisy regions. This is also reflected in the multiclass patient diagnosis task, where SMOTE did not impact XGBoost performance and decreased the performance of the neural network.    

The addition of cost-sensitive weights improved sensitivities of both baseline models for COVID-19 prediction. This effect was especially noticeable with the neural network, as it achieved a much higher mean sensitivity compared to its baseline (mean sensitivity of 0.703 [CI range 0.539-0.785], from 0.236 [CI range 0.071-0.388]). This was also the case for the multiclass diagnosis task, as the sensitivity of minority classes also improved compared to respective baselines (however, at the cost of majority class prediction). This is expected, as cost-sensitive weights put more attention to samples from the minority class, allowing the backpropagation algorithm to assign weights to classification errors in proportion to the class imbalance.

However, in general, reinforcement learning achieved the highest (or second highest) G-mean scores across test sets, except for PUH when threshold adjustment was used. This may be related to the RL method also having poorer overall classification performance on PUH, possibly owing to site-specific differences (e.g. differences in protocols or methods used to collect and process data). RL also achieved the highest (or second highest) F-scores when threshold adjustment was applied. When no threshold adjustment was applied, F scores were not as high compared to other models that had much lower sensitivity (<0.61), but very high specificity (>0.93). This is because these models generally had fewer false positives, which results in a higher F score (as shown in Eq. 12). 

For the multiclass diagnosis task, we found that RL achieved the strongest performance in terms of classification performance and balanced performance (for both individual minority class scores and mean scores across all classes). However, all models performed achieved the lowest sensitivities for predicting acute renal events (which included electrolyte imbalances and renal failure). This may be because none of the markers typically used to diagnose these conditions, such a blood tests, urinalysis, urine output, swelling, or imaging studies, were included as input features to the model. The features available, such as blood pressure, heart rate, and oxygen saturation, primarily reflect the status of the cardiovascular and respiratory systems - as well as systemic function - which could explain why RL performed better in predicting these classes. We also note that the disease categories were neither mutually exclusive nor collectively exhaustive, and that a given patient might have several concurrent diagnosis across labels.  Thus, future studies should consider adding more input features, including laboratory tests and imaging studies, and group diagnoses in a way that accounts for the fact that patient could be assigned to multiple labels. Additionally, multiclass tasks often require more data, and thus, there may not have been sufficient data to confidently differentiate between all classes, especially for this kind of challenging clinical task.

\end{document}